\documentclass{article} 
\usepackage{iclr2027_conference,times}

\usepackage{amsmath,amsfonts,bm}

\def\eqref#1{equation~\ref{#1}}

\def\1{\bm{1}}

\DeclareMathAlphabet{\mathsfit}{\encodingdefault}{\sfdefault}{m}{sl}
\SetMathAlphabet{\mathsfit}{bold}{\encodingdefault}{\sfdefault}{bx}{n}

\usepackage{hyperref}
\usepackage{url}
\usepackage{amsmath,amssymb}
\usepackage{graphicx}
\graphicspath{{figures/}}
\usepackage{booktabs}
\usepackage{multirow}
\usepackage{array}
\usepackage{enumitem}
\usepackage{xcolor}
\usepackage{placeins}
\usepackage{capt-of}
\usepackage{wrapfig}

\newcommand{\wmax}{W_{\max}}
\newcommand{\recon}{\textsc{dasc-wr}}
\newcommand{\norecon}{\textsc{dasc-nr}}
\newcommand{\dasc}{\textsc{dasc}}

\definecolor{gainGreen}{RGB}{0,135,82}

\title{\dasc{}: Decay-Aware State Compression for Hybrid Linear-Attention Serving}
\renewcommand{\iclrpreprintheader}{\dasc{}: Decay-Aware State Compression for Hybrid Linear-Attention Serving}

\author{Yanqi Yu\textsuperscript{1}\thanks{Work done during internships at Meituan.} \quad Pingwei Sun\textsuperscript{2} \quad Jianchao Tan\textsuperscript{2} \quad Tao Zhang\textsuperscript{3} \\[2pt]
\textbf{Yuchen Xie}\textsuperscript{2} \quad \textbf{Xunliang Cai}\textsuperscript{2} \quad \textbf{Yao Liu}\textsuperscript{1}\thanks{Corresponding author.}\\
\textsuperscript{1}East China Normal University \quad \textsuperscript{2}Meituan \quad \textsuperscript{3}South China University of Technology\\
\texttt{yqyu@stu.ecnu.edu.cn, liuyao@cc.ecnu.edu.cn}\\
\texttt{\{sunpingwei,tanjianchao02\}@meituan.com}}

\iclrpreprintcopy 
\hypersetup{pdfauthor={Yanqi Yu, Pingwei Sun, Jianchao Tan, Tao Zhang, Yuchen Xie, Xunliang Cai, Yao Liu},pdfsubject={},pdfkeywords={}}
\begin{document}

\maketitle

\begin{abstract}
Hybrid linear-attention architectures have recently scaled to large open-weight models, offering quality competitive with full attention while substantially reducing key/value (KV) cache growth. However, their in-place recurrent-state updates complicate cache management: prefix reuse requires state checkpoints alongside full-attention KV, while storing state checkpoints in full increases memory pressure, leading to more evictions and repeated prefill.
By analyzing the decay structure of Gated DeltaNet (GDN) and Kimi Delta Attention (KDA), we find that different heads and channels retain prefix information over markedly different timescales, which we term \emph{retention horizons}. This variation suggests substantial compression potential in persistent state checkpoints. Building on this observation, we introduce \emph{Decay-Aware State Compression} (\dasc{}), which derives retention horizons from model weights, packs selected long-horizon units into ragged state checkpoints, and optionally balances their storage across tensor-parallel ranks. On reuse, \dasc{} either zero-fills omitted units or refreshes them from a bounded suffix with additional compute cost.
We evaluate \dasc{} on Kimi-Linear, Qwen3-Next, and Kimi-K3, covering model quality and serving performance. Across the evaluated retrieval and reasoning tasks, conservative configurations maintain quality close to full state checkpoint caching, while suffix refresh recovers accuracy at more aggressive compression. Implemented in SGLang, \dasc{} provides $1.32$--$2.22\times$ state checkpoint compression at representative operating points, reducing mean Time to First Token (TTFT) by $16.1$--$42.0\%$ and mean end-to-end (E2E) latency by $4.9$--$10.2\%$.
\end{abstract}

\section{Introduction}
\label{sec:intro}

Hybrid linear-attention architectures now underpin trillion-parameter open-weight models, including Qwen3.8-2.4T-A95B and the 2.8T-parameter Kimi-K3~\citep{qwen2026qwen38,kimiteam2026kimik3}. These models interleave full-attention layers with linear-attention layers that summarize the prefix in fixed-size recurrent states. This hybrid design retains strong model quality while slowing key/value (KV) cache growth. However, its in-place recurrent-state updates introduce a new challenge for prefix caching.

Unlike full-attention KV, a recurrent state is overwritten as tokens arrive and does not preserve earlier prefix boundaries. Hybrid prefix caching therefore stores recurrent state checkpoints alongside full-attention KV at reusable prefix boundaries~\citep{pan2025marconi,kimiteam2026kimik3}. Because each state checkpoint contains large state matrices for all linear-attention layers, frequent state checkpointing quickly exhausts HBM, while sparse state checkpointing increases replay or repeated prefill. This tension makes it important to determine which state units must be retained at each state checkpoint.

Our analysis reveals a structured answer: recurrent-state units have widely different retention horizons. Many forget distant tokens quickly, while a smaller subset remains persistent. This variation follows the model architecture, appearing across heads in Gated DeltaNet (GDN) and across channels in Kimi Delta Attention (KDA). Moreover, the decay parameters stored in model weights provide an input-independent estimate of these horizons; Figure~\ref{fig:overview} illustrates how they drive state checkpoint selection and reuse. We validate this signal against observed token-dependent decay and state magnitudes. Controlled cache ablations further show why selection must be conservative: recurrent states can be redundant for simple recall yet remain important for multi-key, aggregation, and reasoning tasks.

We translate these findings into \emph{Decay-Aware State Compression} (\dasc{}). Before serving, \dasc{} constructs a static selection plan that retains long-horizon heads for GDN or channels for KDA. At runtime, selected units are flat-packed into ragged state checkpoints with optional tensor-parallel placement balancing; full-attention KV remains unchanged. In SGLang, a two-tier cache combines compressed storage with native slots shared by running states and full cached state checkpoints. On a compressed state checkpoint hit, \norecon{} zero-fills omitted units without additional model computation, whereas \recon{} refreshes them from a bounded suffix at additional compute cost.

We evaluate \dasc{} on Kimi-Linear-48B-A3B-Instruct~\citep{kimiteam2025kimilinear}, Qwen3-Next~\citep{qwen3next2025}, and Kimi-K3~\citep{kimiteam2026kimik3} using retrieval and reasoning tasks together with natural WildChat-1M serving traces~\citep{zhao2024wildchat}. Conservative configurations remain close to dense state checkpoint quality, while suffix refresh recovers accuracy at more aggressive compression. Representative operating points reduce mean TTFT by 42.0\%, 25.6\%, and 16.1\% and mean E2E latency by 10.2\%, 4.9\%, and 6.0\% on the three models, respectively.

Our contributions are:

\begin{itemize}[leftmargin=*,nosep]
    \item We identify the storage pressure of full state checkpoints in hybrid-model prefix caching and characterize decay heterogeneity in GDN and KDA, deriving weight-based retention horizons and validating their ordering against token-dependent decay across three models.

    \item We design and implement \dasc{} in SGLang, integrating architecture-aware selection, ragged storage, optional TP-balanced placement, and two-tier state checkpoint management. Compressed state checkpoints use zero-fill by default; optional bounded-suffix refresh provides accuracy recovery at additional compute cost.

    \item Across Kimi-Linear, Qwen3-Next, and Kimi-K3, conservative configurations maintain quality close to full state checkpoints. Representative operating points achieve $1.32$--$2.22\times$ state checkpoint compression, reducing mean TTFT by $16.1$--$42.0\%$ and mean E2E latency by $4.9$--$10.2\%$.
\end{itemize}

\section{Related Work}
\label{sec:related}

\paragraph{Linear attention and hybrid architectures.} Linear attention and state-space models replace the context-length-dependent KV cache of softmax attention with a fixed-size recurrent state \citep{katharopoulos2020linear,gu2023mamba,dao2024mamba2}. Delta-rule and gated variants add content-dependent updates and adaptive decay \citep{yang2024deltanet,yang2024gateddeltanet}, while hybrid models interleave recurrent and full-attention layers to preserve direct access to token-level KV representations \citep{lieber2024jamba,ren2024samba,kimiteam2025kimilinear}. Unlike architecture and training work, we study how the resulting recurrent states should be represented in a serving-time prefix cache.

\paragraph{Prefix caching and full-attention KV management.} SGLang's RadixAttention organizes shared prefixes in a radix tree, while vLLM's PagedAttention provides block-grained KV allocation and reuse \citep{zheng2024sglang,kwon2023vllm}. Hierarchical systems extend reusable KV across GPU and host memory~\citep{gao2024cost,jin2024ragcache}; quantization, eviction, and sparsification further reduce token-indexed KV storage \citep{hooper2024kvquant,liu2024kivi,zhang2023h2o,xiao2024streamingllm,li2024snapkv}. These methods are complementary to \dasc{}, which leaves full-attention KV unchanged and compresses the recurrent state checkpoint stored alongside it.

\paragraph{Hybrid recurrent-state prefix caching and replay.} Processing a prefix through a recurrent layer produces one boundary state summarizing all preceding tokens, so reuse requires storing that state or recomputing it. Kimi-K3 persists KDA states for prefix reuse~\citep{kimiteam2026kimik3}; Marconi instead improves which hybrid prefix entries are admitted and evicted based on reuse and compute--memory utility \citep{pan2025marconi}. ReplaySSM caches recent SSM inputs and reconstructs states on demand, targeting decode-time state traffic and rollback~\citep{dao2026replayssm}. Q-Mamba quantizes recurrent state caches alongside weights and activations~\citep{tianqi2025qmamba}. Whereas quantization reduces bits per value, \dasc{} selects which units to store, complementing both precision reduction and entry-management policies.

\section{Retention Heterogeneity in Hybrid Recurrent States}
\label{sec:thesis}

This section first uses a controlled cache ablation to identify task-specific redundancy in recurrent state checkpoints. The decay structures of Qwen3-Next's GDN and the KDA modules in Kimi-Linear and Kimi-K3 then motivate a static, serving-input-independent metric for estimating retention horizons. Empirical validation supports its use as a ranking heuristic for state selection, providing the basis for \dasc{}.

\subsection{Recurrent states in hybrid prefix caches}
\label{sec:prelim}

A gated delta-rule layer summarizes the processed sequence in a recurrent state $S_t\in\mathbb{R}^{d_v\times d_k}$, which a query reads as $o_t=S_tq_t$. KDA updates this state as
\begin{equation}
\label{eq:state-full}
S_t \;=\; S_{t-1}\,\mathrm{Diag}(\alpha_t)\,\bigl(I-\beta_t k_t k_t^{\top}\bigr)
\;+\; \beta_t\, v_t k_t^{\top},
\qquad \alpha_t \in (0,1)^{d_k},
\end{equation}
where $\alpha_t$ controls input-dependent forgetting and $\beta_t$ the update strength~\citep{kimiteam2025kimilinear}. The three operations attenuate the old state, remove content aligned with $k_t$, and write the new key--value association. GDN applies gated decay per head~\citep{yang2024gateddeltanet}, whereas KDA exposes channel-wise decay within each head.

Unlike full attention, a delta-rule layer carries only this fixed-size state forward. Reusing prefix $p$ therefore requires both exact full-attention KV and each recurrent layer's boundary state $S^{(l)}(p)$. In prefix caching, each reusable prefix requires its own \emph{recurrent state checkpoint}; if the state checkpoint is not cached, the state must be recomputed through prefill.

\subsection{Recurrent state checkpoints contain task-specific redundancy}
\label{sec:ablation}

We probe these two cache components on RULER NIAH-S1 using five needle depths and $N=100$ examples per depth. For Qwen3-Next-80B-A3B-Instruct and Kimi-Linear-48B-A3B-Instruct, we cache the shared prefix and zero its full-attention KV, recurrent state checkpoints, both, or neither before reuse. Every intervention reuses the same fully cached prefix, so any accuracy difference comes from the zeroed cache component rather than from differences in cache hits.

\begin{table*}[!t]
\caption{Evidence for recurrent-state compression opportunity. \textbf{(a)} Cached-state intervention on RULER NIAH-S1. \textbf{(b)} Within-layer Spearman correlations: $H_s$ is the static horizon, $H_e$ the empirical horizon, and $M_S$ the state norm. Values are median [IQR].}
\label{tab:compression-evidence}
\label{tab:ablation}
\label{tab:horizon-validation}
\centering
\begin{minipage}[t]{0.49\textwidth}
\centering
\textbf{(a) Cached-state intervention}\par\vspace{2pt}
\scriptsize
\setlength{\tabcolsep}{2.5pt}
\renewcommand{\arraystretch}{1.62}
\begin{tabular*}{\linewidth}{@{\extracolsep{\fill}} l c c l}
\toprule
\bf Intervention & \bf Qwen3-Next & \bf Kimi-Linear & \bf Observation \\
\midrule
\textsc{baseline} & $1.000$ & $1.000$ & reference \\
\textsc{zero-full-kv} & $0.000$ & $0.000$ & retrieval fails \\
\textsc{zero-linear} & $0.990$ & $1.000$ & $\le 1$-point change \\
\textsc{zero-both} & $0.000$ & $0.000$ & sanity check \\
\bottomrule
\end{tabular*}
\end{minipage}
\hfill
\begin{minipage}[t]{0.47\textwidth}
\centering
\textbf{(b) Static horizons vs. observed signals}\par\vspace{2pt}
\scriptsize
\setlength{\tabcolsep}{2.0pt}
\renewcommand{\arraystretch}{0.98}
\begin{tabular*}{\linewidth}{@{\extracolsep{\fill}} llcc}
\toprule
\bf Model & \bf Pair & \bf Median & \bf IQR \\
\midrule
Qwen3-Next & $H_s,H_e$ & $.801$ & $[.699,.877]$ \\
Qwen3-Next & $H_s,M_S$ & $.632$ & $[.541,.706]$ \\
\midrule
Kimi-Linear & $H_s,H_e$ & $.856$ & $[.843,.878]$ \\
Kimi-Linear & $H_s,M_S$ & $.533$ & $[.462,.602]$ \\
\midrule
Kimi-K3 & $H_s,H_e$ & $.895$ & $[.867,.941]$ \\
Kimi-K3 & $H_s,M_S$ & $.443$ & $[.363,.513]$ \\
\bottomrule
\end{tabular*}
\end{minipage}
\end{table*}

Removing full-attention KV reduces accuracy to zero, whereas removing recurrent state checkpoints changes it by at most one point (Table~\ref{tab:ablation}(a)). This probe is consistent with retrieval being carried by full-attention KV despite recurrent-state removal. It motivates testing compression, but does not establish general state redundancy; harder tasks in \S\ref{sec:exp-cap} assess its scope.

\subsection{Static horizons enable input-independent selection}
\label{sec:granularity}

The intervention suggests task-specific tolerance to state omission, but does not identify which units to retain. Although the input-dependent decay gate provides a natural retention signal, collecting it for every prompt would make the selection request-dependent. An offline serving plan instead requires a fixed metric that does not depend on serving requests.

To build this metric, DASC evaluates each architecture's static log-decay. Qwen3-Next's GDN and Kimi-Linear's KDA use the standard softplus gate, while Kimi-K3's KDA uses a bounded safe gate~\citep{kimiteam2025kimilinear,kimiteam2026kimik3}:
\begin{equation}
\label{eq:static-decay}
\begin{aligned}
g_s^{\mathrm{std}}(u)
  &= -\exp(A_{\log,u})\,\mathrm{softplus}(a_s+d_u),\\
g_s^{\mathrm{K3}}(u)
  &= \ell\,\sigma\!\left(\exp(A_{\log,u})(a_s+d_u)\right),
     \quad \ell=-5 .
\end{aligned}
\end{equation}
where $A_{\log,u}$ is a learned log-scale parameter, $d_u$ the learned decay bias, and $\sigma(x)=(1+e^{-x})^{-1}$ the sigmoid function. The constant $\ell=-5$ is Kimi-K3's safe-gate lower bound. By contrast, DASC replaces the input-dependent gate input $a_{t,u}$ with a fixed representative value $a_s$, shared across all recurrent layers and units of a model: $a_s=-0.3$ for Qwen3-Next and Kimi-Linear, and $a_s=0.8$ for Kimi-K3. With $\alpha_s=\exp(g_s)$ and $g_s<0$, we define the \emph{static retention horizon} as
\begin{equation}
\label{eq:tau}
H_s=\frac{\ln\epsilon}{g_s}, \qquad \epsilon=10^{-3}.
\end{equation}
For Qwen3-Next and Kimi-Linear, setting $a_s=-0.3$ yields
slower static decay and longer retention horizons than $a_s=0$,
biasing the selector toward retaining more units. For Kimi-K3,
DASC instead evaluates its native bounded safe gate using the
calibrated representative $a_s=0.8$.

\begin{figure*}[!t]
\centering
\includegraphics[width=0.94\textwidth]{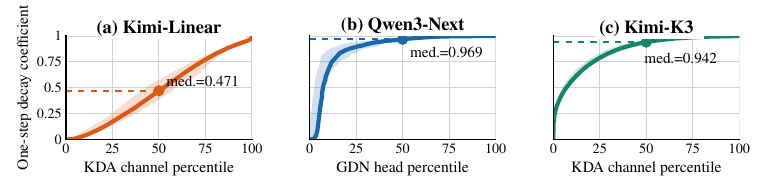}
\vspace{-0.65em}
\caption{Weight-only one-step decay profiles for Kimi-Linear KDA, Qwen3-Next GDN, and Kimi-K3 KDA.}
\label{fig:decay-profile}
\vspace{-0.55em}
\end{figure*}

Figure~\ref{fig:decay-profile} profiles static one-step decay across the three models. Kimi-Linear exhibits broad channel-wise decay (median $\alpha=0.471$), whereas Qwen3-Next GDN heads and Kimi-K3 KDA channels are predominantly slow-decaying (medians $0.969$ and $0.942$). GDN assigns one horizon per head, while KDA resolves differences among channels within each head. At a 16-token cutoff, $60.8\%$ of Kimi-Linear KDA layer--head pairs contain both retained and omitted channels; at a 64-token cutoff, the corresponding fraction is $98.3\%$ for Kimi-K3. This within-head variation motivates channel-wise KDA selection. Across layers, unequal retained-unit counts motivate ragged storage; Appendix Figures~\ref{fig:horizon-gdn}--\ref{fig:k3-retained-fraction} show the model-specific structure and statistics. These weight-derived, request-independent profiles characterize storage heterogeneity, not the importance of omitted units.

\subsection{Weight-derived horizons track observed decay}
\label{sec:horizon-evidence}

We assess how static horizons correlate with empirical horizons and observed state magnitudes across all three models. Qwen3-Next and Kimi-Linear use 32 length-stratified \href{https://huggingface.co/datasets/anon8231489123/ShareGPT_Vicuna_unfiltered}{ShareGPT} prefixes, while Kimi-K3 uses 16 held-out prefixes spanning 128--2,048 tokens. Static horizons follow the model-specific gate parameterizations and representative values defined in Section~\ref{sec:granularity}. Appendix~\ref{app:horizon-validation} details the protocols.

For prefix $p$, recurrent layer $l$, and state unit $u$ (a GDN head or a KDA key channel), let $\mathcal{T}_p$ denote the non-padding token positions and $g_{p,t,l,u}$ the observed log-decay. The \emph{empirical horizon} is
\begin{equation}
\label{eq:empirical-horizon}
\bar g^{(e)}_{p,l,u}=\frac{1}{|\mathcal{T}_p|}\sum_{t\in\mathcal{T}_p}g_{p,t,l,u},
\qquad H^{(e)}_{p,l,u}=\frac{\ln\epsilon}{\bar g^{(e)}_{p,l,u}}.
\end{equation}
We also measure state magnitude at the final prefix position. Suppressing prefix, layer, and time indices, let $S_h$ denote a head state and $S_{h,k}$ a state channel:
\begin{equation}
\label{eq:validation-signals}
M_S(u)=
\begin{cases}
\|S_h\|_F, & u=h \text{ (GDN)},\\
\|S_{h,k}\|_2, & u=(h,k) \text{ (KDA)}.
\end{cases}
\end{equation}

For each prefix--layer pair, we compute Spearman correlations across units between $H_s$ and each of $H_e$ and $M_S$. Table~\ref{tab:horizon-validation}(b) reports medians and IQRs. The median horizon correlations are $0.801$, $0.856$, and $0.895$ for Qwen3-Next, Kimi-Linear, and Kimi-K3; the corresponding state-magnitude correlations are $0.632$, $0.533$, and $0.443$. These results support static horizons as a ranking heuristic associated with observed state activity, rather than a complete importance measure. Rank agreement does not establish agreement at a fixed horizon cutoff; the downstream quality experiments evaluate the effect of the resulting state selection.

\FloatBarrier

\section{Decay-Aware State Compression}
\label{sec:idea1}

\begin{figure*}[!b]
\centering
\includegraphics[width=\linewidth]{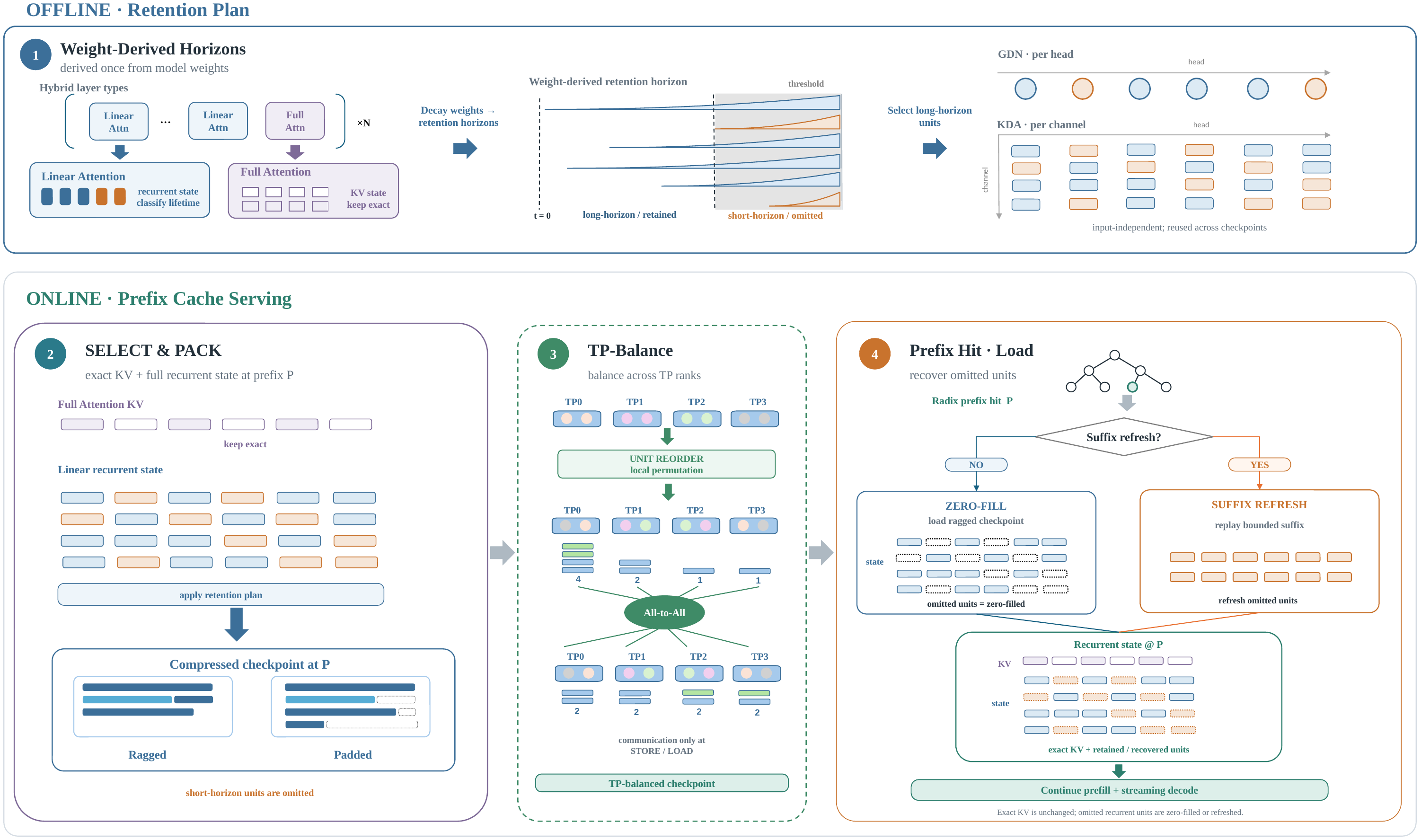}
\caption{\textbf{Overview of \dasc{}.} Weight-derived retention horizons select long-horizon state units, which are packed into ragged state checkpoints with optional TP-balanced placement. On reuse, omitted units are either zero-filled or refreshed from a bounded suffix.}
\label{fig:overview}
\vspace{-1.5em}
\end{figure*}

Using the static horizons $H_s$ from \S\ref{sec:thesis}, \dasc{} uses an input-independent mask to retain long-horizon state units and optionally refreshes omitted units on reuse. Figure~\ref{fig:overview} summarizes the store and load paths.

\subsection{Decay-aware selection and storage}
\label{sec:tau}

\dasc{} constructs a fixed selection mask using a retention-horizon cutoff $\wmax$, measured in tokens. A unit is \emph{global} and retained when $H_s(u)>\wmax$; otherwise, it is \emph{local} and omitted from the persistent state checkpoint. Selection follows the architecture's decay granularity: one complete head state for GDN and one key channel $(h,k)$, corresponding to $S^{(l)}_{h,:,k}$, for KDA. After fixing the model-level representative $a_s$, the mask is constructed once offline and remains independent of serving requests.

Let $L$ be the number of recurrent linear-attention layers, excluding full-attention layers, and let $n_l$ and $r_l$ be the total and retained state-unit counts in layer $l$, respectively. A unit is one head in GDN and one $(\text{head},k)$ channel in KDA. Units have equal cost within the temporal-state tensor, so ragged storage keeps $\sum_l r_l$ temporal units instead of $\sum_l n_l$. Recurrent state checkpoints use SGLang's default mixed precision: the temporal state is FP32 and the convolution state is BF16; BF16 model execution is a separate setting. Let $B_{\mathrm{temp}}$ and $B_{\mathrm{conv}}$ denote their physical bytes. The temporal-only and full state checkpoint compression ratios are
\begin{equation}
\label{eq:capacity}
C_{\mathrm{temp}}^{\mathrm{rag}}=\frac{\sum_{l=1}^{L}n_l}{\sum_{l=1}^{L}r_l},
\qquad
C_{\mathrm{ckpt}}^{\mathrm{rag}}=\frac{B_{\mathrm{temp}}^{\mathrm{dense}}+B_{\mathrm{conv}}}{B_{\mathrm{temp}}^{\mathrm{rag}}+B_{\mathrm{conv}}}.
\end{equation}
At fixed state checkpoint memory, $C_{\mathrm{ckpt}}$ is the ideal capacity multiplier, before slot rounding and allocator reservations; actual capacity also depends on per-rank allocation (Appendix~\ref{app:stats}). A \textbf{padded} temporal layout reserves the largest retained count $R_{\max}=\max_l r_l$ for every layer, even though only $r_l$ entries are valid in layer $l$, giving $C_{\mathrm{temp}}^{\mathrm{pad}}=(\sum_l n_l)/(L R_{\max})$. The \textbf{ragged} layout flat-packs each layer's retained units with offsets. With $\overline{r}=L^{-1}\sum_l r_l$, its layout gain over padded storage is $R_{\max}/\overline{r}$. Reported capacity ratios include the uncompressed BF16 convolution state and the per-rank TP bottleneck (Appendix~\ref{app:layouts}); the gain is large for KDA's imbalanced retained counts and modest for GDN's more uniform profile.

\subsection{Load policies}
\label{sec:load}

On a compressed state checkpoint hit, \dasc{} loads the retained global units exactly. \textbf{\norecon{}} (the default) leaves omitted local units at zero. \textbf{\recon{}} instead replays a bounded suffix from a zero-initialized scratch recurrent state and copies only the refreshed local units into the active state; the stored global state checkpoint and cached full-attention KV remain unchanged. Let $\mathcal{L}_{\mathrm{local}}$ denote the omitted units and $n_p$ the cached prefix length. For $\wmax\ge4$, define $\mathrm{next\_pow2}(x)=2^{\lceil\log_2x\rceil}$ for $x>0$, mapping zero to 4. The longest per-unit window is capped by the available prefix; if no units are omitted, $W_{\mathrm{replay}}=0$ and refresh is skipped. Otherwise,
\begin{equation}
\label{eq:replay-window}
\resizebox{0.68\columnwidth}{!}{$\displaystyle
W_u = \mathrm{clamp}(\mathrm{next\_pow2}(H_s(u)),4,\wmax)
\qquad
W_{\mathrm{replay}} = \min\!\left(n_p,\max_{u\in\mathcal{L}_{\mathrm{local}}} W_u\right).
$}
\end{equation}
Increasing $\wmax$ omits more units and improves state checkpoint compression, but can increase \norecon{} approximation error and \recon{} replay cost. During scratch replay, all recurrent units, including global units, start from zero. Missing pre-window history therefore affects intermediate activations across layers, not only the local-state initial condition. Only refreshed local units are copied back; retained active global units remain exact. Thus \recon{} is approximate recovery, not an exact reproduction of dense prefix computation.

\subsection{TP-balanced placement}

With uneven rank payloads $U_r$, TP state checkpoint capacity is limited by $\max_r U_r$. \dasc{} optionally permutes head bundles and all coupled tensors at model loading. Independently, storage balancing redistributes packed units across $T$ ranks with target slot width $C=\lceil\sum_r U_r/T\rceil$. A variable-split all-to-all transfers only remotely owned units during \textsc{Store}; \textsc{Load} reverses the mapping to restore the compute-local layout. Both stages are exact reindexings; communication occurs only at state checkpoint boundaries, while recurrence remains rank-local. Storage balancing also works without permutation. Appendix~\ref{app:tp-algorithm} gives the assignment algorithms and compatibility constraints.

\subsection{SGLang integration}
\label{sec:integration}

\begin{figure}[!htbp]
  \centering
  \includegraphics[width=\linewidth]{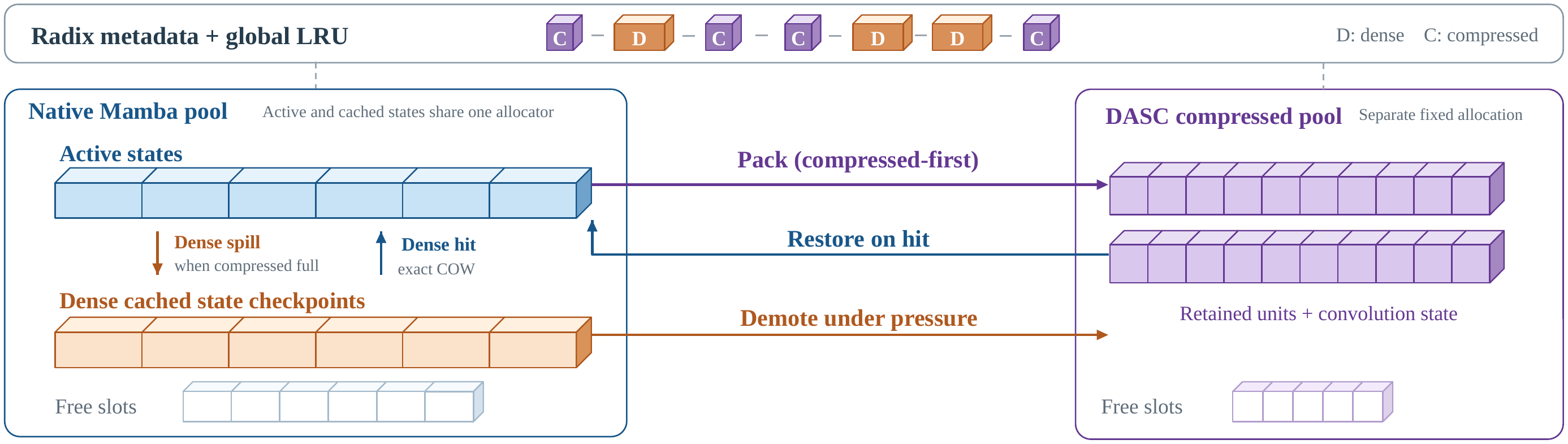}
  \caption{Two-tier state checkpoint cache in SGLang.}
  \label{fig:pool-architecture}
\end{figure}

Figure~\ref{fig:pool-architecture} illustrates DASC's two-tier state checkpoint cache in SGLang. The native Mamba pool shares a slot allocator between running states and full cached state checkpoints, while a separately allocated compressed pool stores retained recurrent units and uncompressed convolution states. Radix metadata tracks each cached prefix's tier, slot, and lock count, and a global LRU maintains a single recency order across both cached representations. KV allocation remains separate. Appendix~\ref{app:pool-management} details insertion, cache reuse, eviction, and asynchronous demotion.
\FloatBarrier

\section{Experiments}
\label{sec:exp}

We evaluate \dasc{} for quality, matched-HBM WildChat serving, and the accuracy--latency trade-off of suffix refresh.

\subsection{Setup}
\label{sec:exp-setup}

\paragraph{Models and hardware.} We evaluate Qwen3-Next-80B-A3B-Instruct, which contains 36 GDN and 12 full-attention layers~\citep{qwen3next2025}, and Kimi-Linear-48B-A3B-Instruct, which contains 20 KDA and 7 full-attention layers~\citep{kimiteam2025kimilinear}. The additional serving and five-task quality study includes Kimi-K3~\citep{kimiteam2026kimik3}. We use Kimi-Linear, Qwen3-Next, and Kimi-K3 for model-level results, and GDN/KDA only for their recurrent architectures. Qwen3-Next exposes head-wise GDN decay, whereas the two Kimi models expose channel-wise KDA decay. Kimi-Linear and Qwen3-Next run on eight Hopper-architecture GPUs with TP8.

\paragraph{Benchmarks.} Long-context quality is evaluated on all 13 RULER subtasks at $4$k, $8$k, and $16$k context lengths, with $N=30$ instances per subtask--length setting~\citep{hsieh2024ruler}. The end-to-end suite covers mathematical reasoning with the MathArena releases of AIME~2026 Parts~I and II and HMMT February~2026~\citep{dekoninck2026matharena}, together with IMO-AnswerBench~\citep{imoanswerbench2025}; general reasoning with GPQA-Diamond~\citep{rein2024gpqa} and MMLU-Pro~\citep{wang2024mmlu}; and conversational memory with LoCoMo~\citep{maharana2024locomo}. Serving performance uses natural conversations from WildChat-1M~\citep{zhao2024wildchat}.

\paragraph{Implementation and evaluation protocol.} We implement DASC in SGLang~\citep{zheng2024sglang}. For Kimi-Linear and Qwen3-Next, model weights and execution are unquantized BF16. Dense denotes the baseline that caches full recurrent state checkpoints without DASC compression, using FP32 temporal states and BF16 convolution states. Unless noted, \norecon{} and ragged storage are used, and compression is relative to the dense mixed-precision state checkpoint, excluding unchanged full-attention KV. Table~\ref{tab:reasoning} and RULER use a strict warmup--replay protocol: an unscored warm request populates each prefix state checkpoint, and only replay outputs and pooled replay token-hit rates are reported. Replays are averaged within each unique problem or instance, which remains the statistical unit; paired arms share prompts, cache-population order, seeds, and slot counts. DASC and Random accuracy runs cache state checkpoints only in the compressed pool, with dense spill disabled; native dense slots serve active computation only. WildChat serving instead enables the two-tier cache with dense spill (Appendix~\ref{app:pool-management}) and reports all-request means over a continuous trace. Appendix~\ref{app:stats} gives task-specific counts and controls.
\subsection{Quality under recurrent-state compression}
\label{sec:exp-cap}

RULER uses three measured replay rounds for each of $N=30$ unique instances per subtask--length--arm setting. The NIAH-S1 diagnostic is saturated: dense and \norecon{} both score $1.000$ for both models at every tested $\wmax$. We therefore report the broader RULER retrieval, extraction, QA, and tracking suite. Appendix Tables~\ref{tab:ruler-kda-extra} and~\ref{tab:ruler-gdn-full} report the complete Kimi-Linear and Qwen3-Next sweeps.

\begin{table*}[t]
\caption{Replay-only end-task quality. \dasc{} entries are point estimates; Random is mean$\pm$sample SD across the same five fixed-mask seeds for all three models. LoCoMo reports token F1; other rows report accuracy. Hit is the replay token-hit rate pooled across all compared arms in each row.}
\label{tab:reasoning}
\centering
\scriptsize
\setlength{\tabcolsep}{2.2pt}
\renewcommand{\arraystretch}{0.90}
\resizebox{\textwidth}{!}{%
\begin{tabular}{llcc!{\vrule width 0.3pt}cc!{\vrule width 0.3pt}cc!{\vrule width 0.3pt}cc}
\toprule
\multicolumn{4}{c}{} & \multicolumn{6}{c}{\bf Ragged \norecon{}, by $\wmax$} \\
\cmidrule(lr){5-10}
\bf Model & \bf Benchmark & \bf Hit & \bf Dense & \multicolumn{2}{c!{\vrule width 0.3pt}}{$\wmax=16$} & \multicolumn{2}{c!{\vrule width 0.3pt}}{$\wmax=64$} & \multicolumn{2}{c}{$\wmax=128$} \\
\cmidrule(lr){5-6}\cmidrule(lr){7-8}\cmidrule(lr){9-10}
& & & & \bf \dasc{} & \bf Random & \bf \dasc{} & \bf Random & \bf \dasc{} & \bf Random \\
\midrule
\multirow{6}{*}{\rotatebox[origin=c]{90}{Kimi-Linear}} & AIME 2026 & 0.762 & 0.6065 & 0.6032 & $0.5634\pm0.0356$ & 0.5398 & $0.5153\pm0.0329$ & $0.5699$ & $0.4744\pm0.0252$ \\
 & HMMT 2026 Feb & 0.705 & 0.3754 & $0.3861$ & $0.3423\pm0.0160$ & 0.3343 & $0.3062\pm0.0167$ & 0.3099 & $0.2741\pm0.0180$ \\
 & IMOAnswerBench & 0.546 & 0.2783 & 0.2675 & $0.2663\pm0.0063$ & 0.2717 & $0.2568\pm0.0066$ & 0.2642 & $0.2455\pm0.0066$ \\
 & GPQA-Diamond & 0.716 & 0.6185 & $0.6131$ & $0.5749\pm0.0189$ & 0.5579 & $0.5422\pm0.0083$ & $0.5704$ & $0.5309\pm0.0111$ \\
 & MMLU-Pro & 0.752 & 0.6750 & $0.6732$ & $0.6550\pm0.0046$ & $0.6458$ & $0.6343\pm0.0015$ & $0.6386$ & $0.6236\pm0.0028$ \\
 & LoCoMo (F1) & 0.993 & 0.4995 & 0.4986 & $0.4883\pm0.0093$ & 0.4899 & $0.4640\pm0.0070$ & 0.4824 & $0.4568\pm0.0141$ \\
\midrule
\multirow{6}{*}{\rotatebox[origin=c]{90}{Qwen3-Next}} & AIME 2026 & 0.621 & 0.6763 & 0.6871 & $0.6735\pm0.0157$ & 0.6763 & $0.6813\pm0.0173$ & 0.6753 & $0.6802\pm0.0095$ \\
 & HMMT 2026 Feb & 0.578 & 0.3881 & 0.3861 & $0.3867\pm0.0072$ & 0.3803 & $0.3975\pm0.0167$ & 0.3949 & $0.3871\pm0.0078$ \\
 & IMOAnswerBench & 0.499 & 0.3375 & 0.3400 & $0.3440\pm0.0090$ & 0.3383 & $0.3472\pm0.0125$ & 0.3417 & $0.3425\pm0.0072$ \\
 & GPQA-Diamond & 0.819 & 0.6960 & 0.7013 & $0.7013\pm0.0089$ & 0.7051 & $0.6998\pm0.0064$ & 0.7044 & $0.7034\pm0.0047$ \\
 & MMLU-Pro & 0.868 & 0.7875 & 0.7882 & $0.7857\pm0.0082$ & 0.7863 & $0.7828\pm0.0063$ & 0.7845 & $0.7815\pm0.0127$ \\ & LoCoMo (F1) & 0.993 & 0.4074 & 0.4158 & $0.4105\pm0.0020$ & 0.4153 & $0.4133\pm0.0033$ & 0.4161 & $0.4101\pm0.0038$ \\
\midrule
\multicolumn{4}{c}{} & \multicolumn{2}{c!{\vrule width 0.3pt}}{$\wmax=64$} & \multicolumn{2}{c!{\vrule width 0.3pt}}{$\wmax=128$} & \multicolumn{2}{c}{$\wmax=512$} \\
\multirow{5}{*}{\rotatebox[origin=c]{90}{Kimi-K3}} & AIME 2026 & 0.8053 & 0.9656 & 0.9688 & $0.9247\pm0.0083$ & 0.9538 & $0.9312\pm0.0145$ & 0.9570 & $0.9206\pm0.0069$ \\
 & HMMT 2026 Feb & 0.7235 & 0.9257 & 0.9257 & $0.8970\pm0.0092$ & 0.9218 & $0.8964\pm0.0115$ & 0.9120 & $0.8952\pm0.0170$ \\
 & IMOAnswerBench & 0.4616 & 0.6417 & 0.6442 & $0.5977\pm0.0073$ & 0.6375 & $0.6025\pm0.0170$ & 0.6150 & $0.5667\pm0.0043$ \\
 & MMLU-Pro & 0.8616 & 0.8809 & 0.8806 & $0.8734\pm0.0037$ & 0.8793 & $0.8651\pm0.0158$ & 0.8753 & $0.8620\pm0.0092$ \\& LoCoMo (F1) & 0.9988 & 0.6672 & 0.6714 & $0.6511\pm0.0185$ & 0.6580 & $0.6479\pm0.0062$ & 0.6594 & $0.6230\pm0.0128$ \\
\bottomrule
\end{tabular}%
}
\end{table*}

At $\wmax=16$, Kimi-Linear scores $0.95/0.95/0.95$ at $4$k/$8$k/$16$k versus dense $0.95/0.96/0.95$; Qwen3-Next scores $0.84/0.83/0.81$ versus $0.84/0.83/0.80$. Thus every displayed aggregate is within $0.01$ of dense. At $\wmax=1024$, Kimi-Linear drops by $0.04/0.06/0.07$, with the largest $16$k losses on MK1, MV, CWE, and QA-S. Qwen3-Next aggregates remain within $0.01$, although MK3 falls from $1.00$ to $0.87$ at $16$k. Section~\ref{sec:recon-cost} evaluates suffix refresh.

\paragraph{End-task quality.} Table~\ref{tab:reasoning} applies the shared warmup--replay protocol to dense state checkpoints, ragged \norecon{}, and count-matched Random. Scores average measured replay generations within each problem, and Hit pools cached and new tokens across all compared arms in each row; arm-specific hit rates are similar. For each layer and $\wmax$, each of five fixed Random masks uniformly retains the same number of units as \dasc{}, holding compression fixed to isolate the benefit of decay-aware ranking. At $\wmax=16$, \dasc{} remains within $1.3$ percentage points of dense on every Kimi-Linear and Qwen3-Next row. On Kimi-Linear, it exceeds the five-mask Random mean in all 15 reasoning comparisons. Qwen3-Next remains within $1.7$ points of dense, with mixed \dasc{}--Random differences. On Kimi-K3, scores across five tasks stay within $1.2$ points of dense at $\wmax=64/128$; at $512$, the largest gap is $2.67$ points on IMOAnswerBench. DASC exceeds the five-mask Random mean in all 15 point-estimate comparisons (Table~\ref{tab:reasoning}).

\noindent
\begin{minipage}[t]{0.60\columnwidth}
\vspace{0pt}
\paragraph{Composition with state quantization.}
Prior work shows that recurrent states can be quantized alongside model weights and activations~\citep{tianqi2025qmamba}. Quantization reduces the bits per retained value, whereas \dasc{} reduces the number of retained units, making the two complementary. We combine $\wmax=16$ channel selection, INT8 state checkpoint storage, and \norecon{} on the five Kimi-Linear end tasks, yielding $8.11\times$ state checkpoint capacity relative to the dense mixed-precision reference. Appendix~\ref{app:int8-dasc-e2e} gives the protocol.
\end{minipage}
\hfill
\begin{minipage}[t]{0.36\columnwidth}
\vspace{-9pt}
\centering
\fontsize{7.2}{7.7}\selectfont
\captionof{table}{Kimi-Linear \dasc{} at $\wmax=16$ (INT8: $8.11\times$).}
\label{tab:int8-dasc-e2e}
\renewcommand{\arraystretch}{0.88}
\setlength{\tabcolsep}{0.45pt}
\begin{tabular*}{\linewidth}{@{\extracolsep{\fill}}lccc@{}}
\toprule
\bf Task & \bf Dense & \shortstack{\bf \dasc{}\\\bf Native} & \shortstack{\bf \dasc{}\\\bf INT8} \\
\midrule
AIME & $0.6065$ & $0.6032$ & $0.6215$ \\
HMMT & $0.3754$ & $0.3861$ & $0.3715$ \\
IMO  & $0.2783$ & $0.2675$ & $0.2775$ \\
GPQA & $0.6185$ & $0.6131$ & $0.6128$ \\
MMLU & $0.6750$ & $0.6732$ & $0.6748$ \\
\bottomrule
\end{tabular*}
\end{minipage}


\begin{figure*}[t]
  \centering
  \includegraphics[width=\textwidth,trim=0 2 0 2,clip]{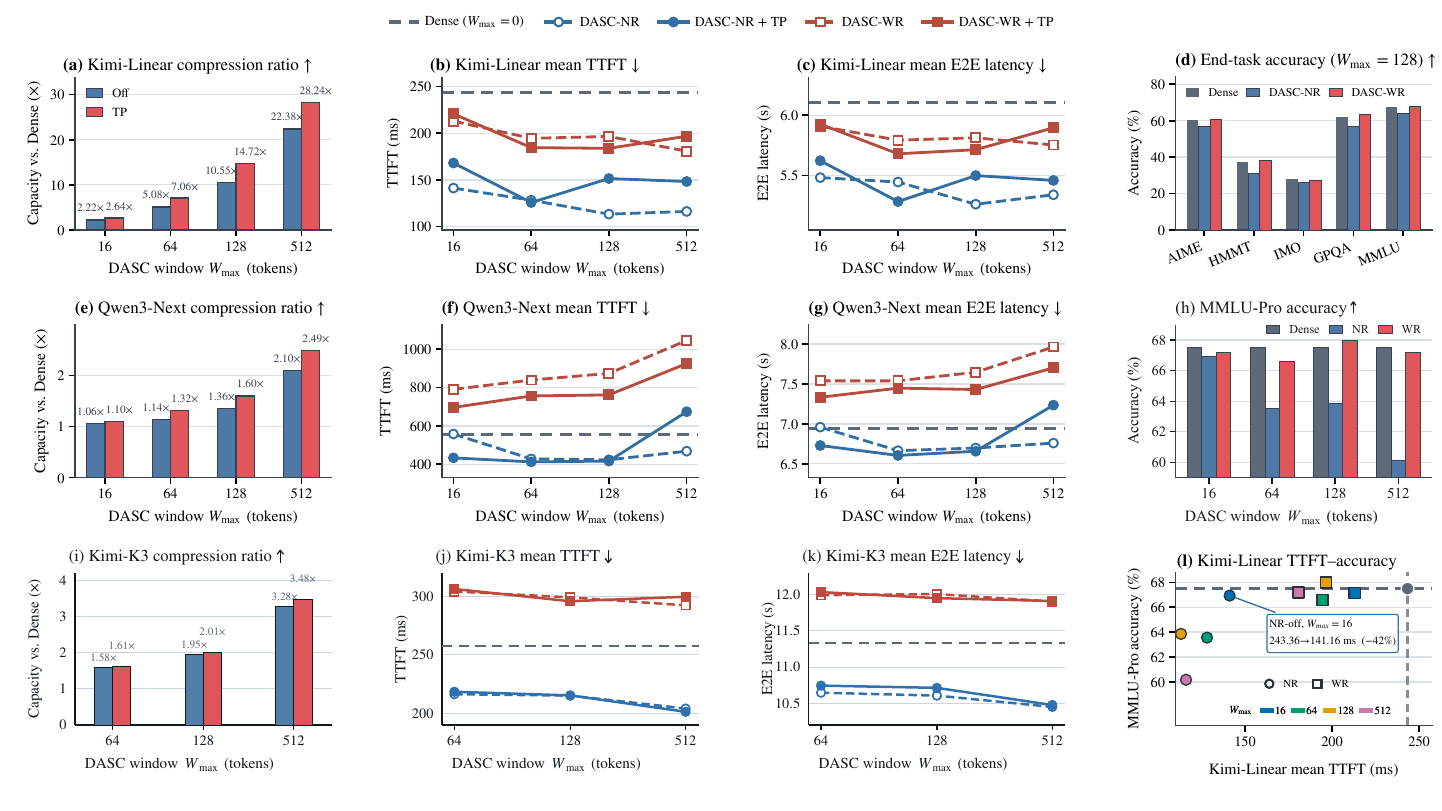}
  \vspace{-0.65em}
  \caption{WildChat-1M serving and accuracy; serving panels report five-seed means. Panels (a--c/e--g/i--k) show state checkpoint capacity, mean TTFT, and mean E2E latency for Kimi-Linear/Qwen3-Next/Kimi-K3 (filled: TP-balanced). Panels (d,h) show Kimi-Linear end-task accuracy; (l) pairs its MMLU-Pro accuracy with TTFT. Arrows indicate preferred directions.}
  \label{fig:kda-gdn-overview}
  \vspace{-0.55em}
\end{figure*}

  \subsection{WildChat serving under matched HBM}
\label{sec:exp-serving}

We evaluate WildChat-1M serving under matched allocated state-memory budgets (Appendix~\ref{app:serving-protocol}). Main serving runs use client concurrency 256 and a server concurrency limit of 32. Figure~\ref{fig:kda-gdn-overview} reports five-seed all-request mean TTFT and E2E latency. ``TP'' enables state checkpoint placement balancing at a fixed TP degree; ``Off'' disables it.

\paragraph{State checkpoint capacity.} Figure~\ref{fig:kda-gdn-overview}(a,e,i) shows that capacity depends on model, window, and placement. With TP-balanced placement, Kimi-Linear reaches $2.64/7.06/14.72/28.24\times$ capacity at $\wmax=16/64/128/512$; Qwen3-Next reaches $1.10/1.32/1.60/2.49\times$; and Kimi-K3 reaches $1.61/2.01/3.48\times$ at $\wmax=64/128/512$. The gap reflects selection granularity and layer imbalance: channel-wise KDA exposes more removable units than head-wise GDN.

\paragraph{Serving performance.} Figure~\ref{fig:kda-gdn-overview}(b,c,f,g,j,k) reports five-seed all-request means, including cache misses and runtime overhead. Selected \norecon{} configurations reduce both TTFT and E2E latency on all three models, while suffix-refresh overhead can offset the capacity benefit. Section~\ref{sec:recon-cost} summarizes the selected configurations and their quality--latency trade-offs.

\subsection{\dasc{}-WR: accuracy--latency trade-off}
\label{sec:recon-cost}

Figure~\ref{fig:kda-gdn-overview}(d,h,l) reports Kimi-Linear accuracy and its trade-off with latency. Panel~(l) pairs Kimi-Linear MMLU-Pro accuracy with WildChat TTFT from separate workloads, providing a cross-workload operating-point comparison.

\paragraph{Accuracy recovery.} At $\wmax=128$, panel~(d) shows that \recon{} improves all five Kimi-Linear end-task estimates over \norecon{} by $0.8$--$7.5$ percentage points. It matches or exceeds Dense on HMMT, GPQA, and MMLU-Pro and remains within $0.7$ points on AIME and IMO. Panel~(h) shows the Kimi-Linear MMLU-Pro sweep: \norecon{} ranges from $60.17$ to $66.94\%$, whereas \recon{} achieves $66.59$--$68.00\%$, close to the $67.50\%$ Dense baseline.

\paragraph{Accuracy--latency trade-off.} We select close-to-Dense \norecon{} operating points: $\wmax=16$ with Off for Kimi-Linear, $\wmax=64$ with TP balancing for Qwen3-Next, and $\wmax=64$ with Off for Kimi-K3. Relative to Dense, these reduce mean TTFT by $42.0/25.6/16.1\%$ and mean E2E latency by $10.2/4.9/6.0\%$, respectively. The corresponding quality differences are $0.56$ points on Kimi-Linear MMLU-Pro, at most $0.92$ points across six Qwen3-Next end tasks, and at most $0.42$ points across five Kimi-K3 end tasks. These cross-workload operating points balance quality and latency: larger windows risk quality loss, while WR adds recovery cost. Keeping the selected Kimi-K3 configuration fixed, we additionally test server concurrency limits of 16 and 8 at client concurrency 256. Under increased queueing pressure, mean TTFT falls by $15.7/11.3\%$ and mean E2E latency by $11.3/9.2\%$, respectively (Appendix Table~\ref{tab:k3-queueing}).



\section{Conclusion}
\label{sec:conclusion}

We presented \dasc{}, combining decay-aware selection, ragged storage, and optional TP-balanced placement for hybrid prefix caches. Conservative configurations retain quality close to Dense. At the selected operating points in Section~\ref{sec:recon-cost}, mean TTFT falls by $42.0/25.6/16.1\%$ and mean E2E latency by $10.2/4.9/6.0\%$ on Kimi-Linear/Qwen3-Next/Kimi-K3, respectively, demonstrating serving benefits across channel-wise KDA and head-wise GDN.
\FloatBarrier

\bibliographystyle{iclr2027_conference} \bibliography{iclr2027_conference}

\clearpage
\appendix
\raggedbottom
\section{Additional retention-horizon analysis}
\label{app:horizon-analysis}

\subsection{Layerwise structure}
\label{app:horizon-structure}

\begin{figure*}[t]
\centering
\includegraphics[width=0.94\textwidth]{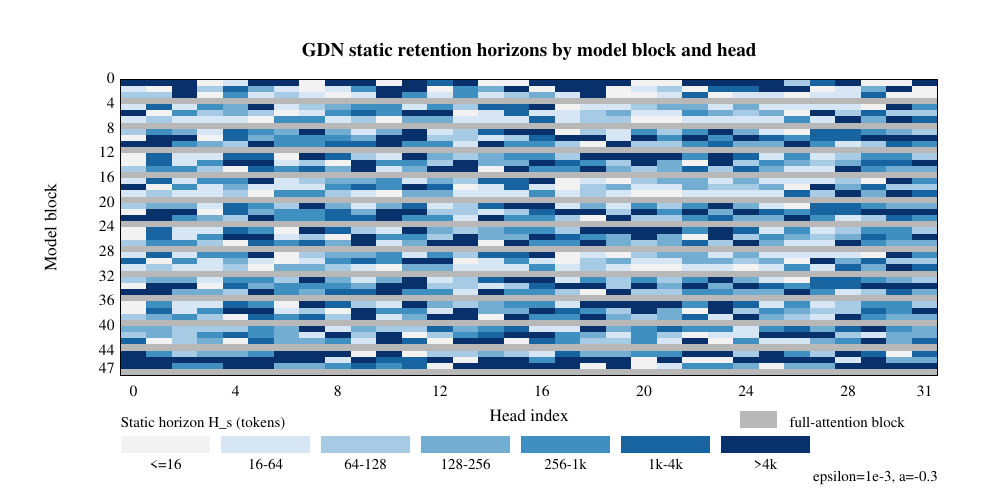}
\caption{Static horizons of all Qwen3-Next GDN heads ($a=-0.3$, $\epsilon=10^{-3}$), in model-block/head
order. Gray rows are full-attention blocks.}
\label{fig:horizon-gdn}
\end{figure*}

\begin{figure*}[t]
\centering
\includegraphics[width=0.94\textwidth]{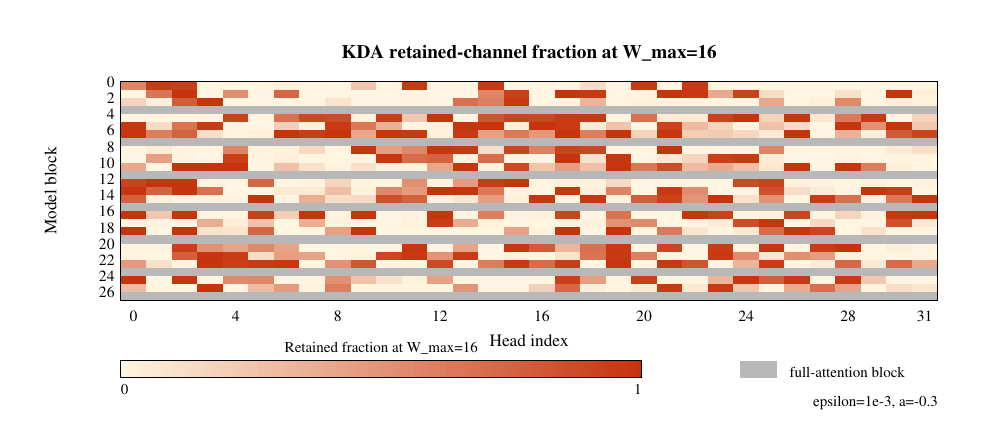}
\caption{Kimi-Linear retained-channel fractions at a 16-token cutoff. Each cell summarizes 128 channels. Across 20 KDA layers and 32 heads, 389 of 640 layer--head pairs ($60.8\%$) contain both retained and omitted channels, 215 omit all channels, and 36 retain all. Layerwise mean retained fractions range from $0.179$ to $0.558$; gray full-attention rows are excluded.}
\label{fig:horizon-kda-retained}
\end{figure*}

\begin{figure*}[t]
\centering
\includegraphics[width=0.94\textwidth]{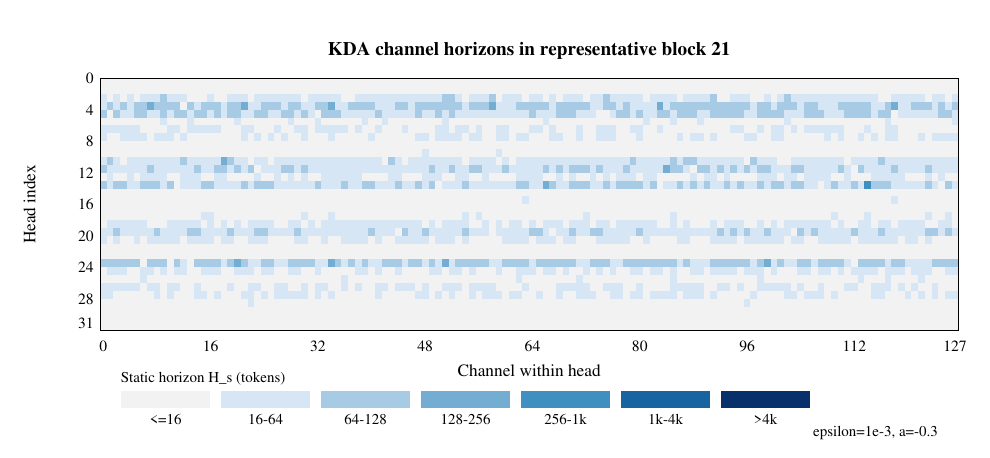}
\caption{Channel horizons in Kimi-Linear block~21, whose retained fraction ($0.375$) is closest to the
layer median ($0.360$). Indices retain architectural order.}
\label{fig:horizon-kda-channels}
\end{figure*}

\begin{figure}[!htbp]
\centering
\includegraphics[width=0.88\textwidth]{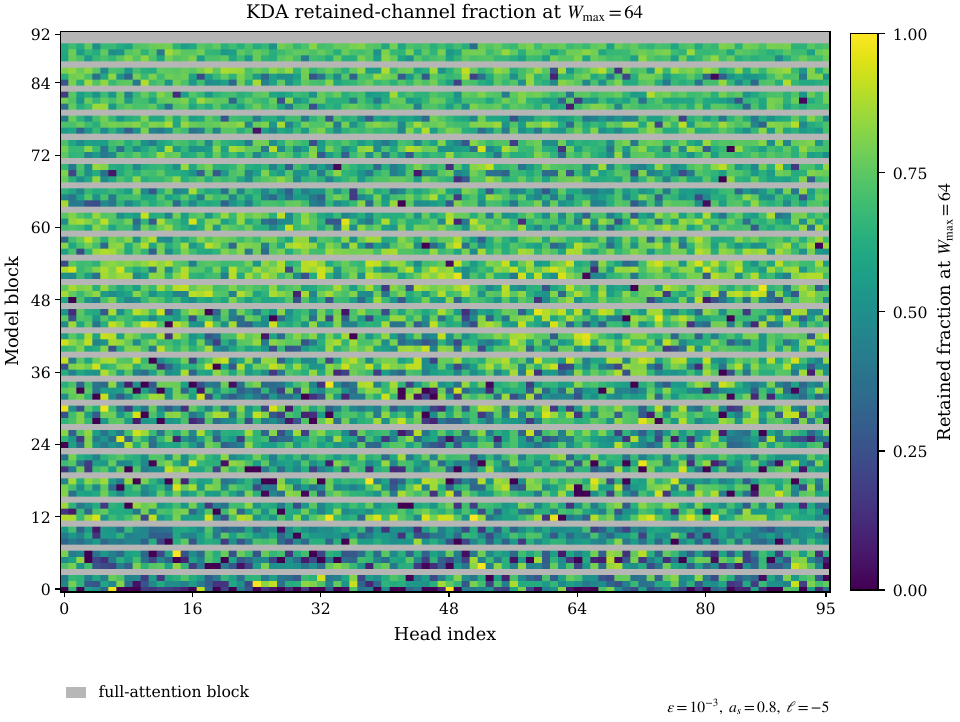}
\caption{Kimi-K3 retained-channel fractions at a 64-token cutoff. Across 93 model blocks, 69 KDA layers contribute 6,624 layer--head pairs; gray full-attention rows are excluded. Of these pairs, 6,512 ($98.3\%$) contain both retained and omitted channels, 108 omit all channels, and 4 retain all. Layerwise mean retained fractions range from $0.267$ to $0.741$.}
\label{fig:k3-retained-fraction}
\end{figure}

Figures~\ref{fig:horizon-gdn}--\ref{fig:k3-retained-fraction} preserve architectural ordering, complementing the pooled decay profiles in Figure~\ref{fig:decay-profile}. Qwen3-Next's GDN has one horizon per head; Kimi-Linear and Kimi-K3 additionally exhibit within-head channel variation. Horizon bins summarize ranges rather than every tested cutoff.

For either KDA model, the retained-channel fraction of layer $l$ and head $h$ at cutoff $w$ is
\[
r_{l,h}(w)=\frac{1}{128}\sum_{k=1}^{128}\mathbf{1}\!\left[H_s(l,h,k)>w\right].
\]
A layer--head pair is mixed when $0<r_{l,h}(w)<1$. The mixed-pair percentage is pooled across all KDA layers and heads, not measured in a single representative layer. Each layer's mean retained fraction averages $r_{l,h}$ over its heads; the ranges in the captions are the minimum and maximum of these layer means. Gray full-attention rows are excluded throughout. Kimi-Linear uses 20 KDA layers with 32 heads each, $a_s=-0.3$, and $w=16$; Kimi-K3 uses 69 KDA layers with 96 heads each, $a_s=0.8$, and $w=64$. Both use $\epsilon=10^{-3}$ and their native gates from Eq.~\ref{eq:static-decay}.

These weight-derived maps are independent of serving requests. Within-head variation motivates channel-wise selection, while unequal layerwise retained counts motivate ragged storage. The maps do not measure state/readout correlation or establish that omitted channels are unimportant; downstream quality experiments evaluate the effect of omission.

\subsection{Static-horizon validation}
\label{app:horizon-validation}
\label{app:k3-safe-gate}

For Qwen3-Next and Kimi-Linear (Table~\ref{tab:horizon-validation}(b)), we deterministically select 32 prefixes per model from the unfiltered cleaned ShareGPT V3 release (\href{https://huggingface.co/datasets/anon8231489123/ShareGPT_Vicuna_unfiltered}{\texttt{anon8231489123/ShareGPT\_Vicuna\_unfiltered}}; 94,145 conversations). Turns are serialized as alternating \texttt{User:} and \texttt{Assistant:} blocks. We process target lengths $16{,}384$, $8{,}192$, $4{,}096$, and $1{,}024$ in descending order so that long conversations are not consumed by shorter tiers, selecting eight prefixes per tier. For each target $T$, we first retain conversations with at least $3.5T$ characters, tokenize the first 500 eligible candidates, sort them by token count, and deterministically choose the eight shortest candidates with at least $T$ tokens. If fewer than eight qualify, we use the longest available candidates, possibly shorter than $T$. Longer sequences are left-truncated to $T$ tokens. Each model independently selects samples with its own tokenizer, so raw conversation sets may differ; source-array indices are released with the analysis output. No random sampling seed is used. Each prefix is processed independently; prefixes are not concatenated. We run a full BF16 forward pass through Qwen3-Next (36 GDN layers) and Kimi-Linear (20 KDA layers), hook the observed token-dependent gates in every linear-attention layer, and execute the corresponding delta-rule recurrence to obtain the recurrent state magnitude at the final prefix position. No analysis prefix is used to construct the deployed mask. Kimi-K3 uses 32 ShareGPT prefixes spanning 128--2,048 tokens, processed independently through all 69 KDA layers (96 heads $\times$ 128 key channels), for 2,208 prefix--layer observations across the full set. The even-indexed 16 prefixes select $a_s=0.8$, and the odd-indexed 16 provide held-out validation; this calibration split and shorter prefix range differ from the other two models. These shorter-prefix measurements characterize gate ordering and final-state activity on that range; they do not establish long-context calibration validity. Longer-context quality is assessed separately, rather than inferred from these correlations.

For all three models, empirical horizons and state magnitudes follow Eqs.~\ref{eq:empirical-horizon} and~\ref{eq:validation-signals}. Static horizons use $\epsilon=10^{-3}$ and each model's native gate in Eq.~\ref{eq:static-decay}, with $a_s=-0.3$ for Qwen3-Next and Kimi-Linear and $a_s=0.8$ for Kimi-K3. We calculate Spearman correlations across units separately for each validation prefix and layer, then report their median and IQR in Table~\ref{tab:horizon-validation}(b). Units from different layers are not pooled.

\section{Buffer layout details}
\label{app:layouts}

\paragraph{Padded (Scheme~1).} Let $R_{\max}=\max_{1\leq l\leq L}r_l$, where $L$ is the number of recurrent linear-attention layers. For GDN, where one retained unit is a complete head state, the pool is $[L,\text{slots},R_{\max},d_v,d_k]$. For KDA, where one retained unit is a $(\text{head},k)$ column, it is $[L,\text{slots},R_{\max},d_v]$. Only the first $r_l$ entries of layer $l$ are valid; the remaining entries are padding. A single wide layer thus pins every layer's temporal allocation, giving $C_{\mathrm{temp}}^{\mathrm{pad}}=(\sum_{l=1}^{L}n_l)/(L R_{\max})$. Full state checkpoint compression additionally includes the BF16 convolution state as in Eq.~\ref{eq:capacity}.

\paragraph{Ragged (Scheme~2).} Each layer contributes only its own $r_l$ retained units, with offsets $o_0=0$ and $o_l=o_{l-1}+r_l$ for $l=1,\ldots,L$. GDN uses $[\text{slots},\sum_{l=1}^{L}r_l,d_v,d_k]$, whereas KDA uses $[\text{slots},\sum_{l=1}^{L}r_l,d_v]$; layer $l$ occupies $[o_{l-1},o_l)$. With $\overline{r}=\tfrac1L\sum_{l=1}^{L}r_l$, the temporal layout achieves $C_{\mathrm{temp}}^{\mathrm{rag}}=(\sum_{l=1}^{L}n_l)/(L\overline{r})$. The retained values match the padded layout, so accuracy is layout-independent. The resulting temporal-layout gain, $C_{\mathrm{temp}}^{\mathrm{rag}}/C_{\mathrm{temp}}^{\mathrm{pad}}=R_{\max}/\overline{r}$, increases with layer imbalance: unequal retained-unit counts create padding overhead that the ragged layout avoids.

\subsection{Deterministic TP balancing}
\label{app:tp-algorithm}

The compute layout and the persistent state checkpoint layout are distinct. Figure~\ref{fig:tp-algorithms} specifies both offline assignment procedures; the resulting maps are reused at runtime. Let $P$ be the TP degree. The cost $c_{l,h}$ of head $h$ in layer $l$ is its retained-channel count for KDA and its retained-head indicator for GDN. Within each store, equal-size units make unit-count and byte-based ordering equivalent. A bundle comprises all coupled tensors of a KDA head ($B=1$), or the $B=H_v/H_k$ adjacent GDN value heads sharing one key head. Its cost is $C_{l,b}=\sum_{h=bB}^{(b+1)B-1}c_{l,h}$. The implementation requires $H_v\bmod B=0$ and $(H_v/B)\bmod P=0$, assigning exactly $K=(H_v/B)/P$ bundles per rank per layer. Cumulative rank loads $L_r$ include all processed layers.

\par\medskip
\noindent\begin{minipage}{\linewidth}
\centering
\setlength{\fboxsep}{6pt}
\fbox{\begin{minipage}[t]{0.46\textwidth}
\footnotesize\raggedright
\textbf{(a) Compute-head permutation}\par\smallskip
\textbf{Input:} bundle costs $C_{l,b}$, TP degree $P$, per-layer quota $K$.\par
\textbf{Output:} bundle owners and loads $L_r$.\par
\begin{enumerate}[leftmargin=1.6em,itemsep=2pt,topsep=3pt]
\item Initialize $L_r=0$ for all ranks.
\item For each layer $l$ in increasing order:
\item[] Set per-rank bundle counts $u_r=0$.
\item[] Sort bundles by $(-C_{l,b},b)$.
\item[] For each bundle $b$, among $u_r<K$, select the lexicographically smallest
\[
(L_r+C_{l,b},\ L_r,\ u_r,\ r).
\]
\item[] Assign $b$ to that rank; increment $u_r$ and add $C_{l,b}$ to $L_r$.
\item Define the load objective
\[
\Phi(\mathbf L)=\bigl(\max_r L_r,\ \sum_r L_r^2\bigr).
\]
\item Scan layers and pairs $b_1<b_2$ in fixed increasing order. Only consider pairs on different ranks in the same layer.
\item Evaluate $\Phi$ after each swap; retain the best strict improvement. Keep the first candidate on equal scores.
\item If no improving swap exists, stop. Otherwise apply it and repeat step 4.
\item Apply the resulting permutation consistently to every head-coupled tensor, including output-projection input columns.
\end{enumerate}
\textbf{Guarantee:} per-layer quotas are preserved; the final assignment is locally optimal for a single within-layer pair swap, not globally optimal.
\end{minipage}}
\hfill
\fbox{\begin{minipage}[t]{0.46\textwidth}
\footnotesize\raggedright
\textbf{(b) State checkpoint storage balancing}\par\smallskip
\textbf{Input:} compute-local unit counts $n_r$ after optional permutation.\par
\textbf{Output:} storage-owner map and transfer splits.\par
\begin{enumerate}[leftmargin=1.6em,itemsep=2pt,topsep=3pt]
\item Compute $N=\sum_r n_r$ and $(q,m)=\operatorname{divmod}(N,P)$.
\item Sort ranks by $(-n_r,r)$; set $t_r=q+1$ for the first $m$ ranks and $t_r=q$ for the rest.
\item Keep the first $\min(n_r,t_r)$ local units on each rank.
\item List excess units by source rank, then local unit index. List deficit positions by destination rank, then position.
\item Pair both lists in order; precompute the destination map and variable split sizes.
\item Allocate $\lceil N/P\rceil$ units per slot on every rank.
\end{enumerate}
\textbf{Runtime store}\par
Compute-local state $\to$ fused pack/partition $\to$ variable-split all-to-all $\to$ scatter into storage slots.\par\smallskip
\textbf{Runtime load}\par
Storage slots $\to$ fused gather/partition $\to$ inverse all-to-all $\to$ scatter into compute-local active state.\par\smallskip
\textbf{Transfer volume:}
\[
M_{\mathrm{remote}}=\sum_r\max(0,n_r-t_r).
\]
For these targets, each excess unit must move and moves once; transfer is minimal for equal-size units. If $M_{\mathrm{remote}}=0$, skip the all-to-all.
\end{minipage}}
\captionof{figure}{Deterministic TP assignment algorithms. (a) Optional compute-side permutation preserves equal bundle counts. (b) Storage balancing changes storage ownership independently of compute ownership. All comparisons and tie-breaking are deterministic.}
\label{fig:tp-algorithms}
\end{minipage}
\par\medskip

\paragraph{Initialization versus runtime communication.} The compute permutation moves Q/K/V and gate projections, recurrent decay parameters, convolution parameters, output-projection input columns, and other coupled axes together. Initialization may use collectives, but subsequent head-separable recurrence uses the locally owned weights and active states. Existing TP collectives remain; permutation adds no token-time collective. Storage balancing leaves compute ownership unchanged, so remotely stored units still require communication at each state checkpoint store/load boundary. Fused/batched operations reduce launches and temporary buffers, not the logically required remote payload. In implementation, \texttt{off} changes neither compute nor cross-rank storage ownership; \texttt{storage} applies storage balancing only; and \texttt{hybrid} adds compute-head permutation before storage balancing. The compute permutation requires unquantized linear weights, so Kimi-K3's MXFP4 packed weights use \texttt{storage} only. This compatibility constraint concerns model weights, not state checkpoint precision. The experimental label \emph{TP} denotes balanced placement collectively, rather than a particular implementation mode.

\subsection{Two-tier state checkpoint management}
\label{app:pool-management}

\paragraph{Allocation and ownership.} The two-tier integration uses a native Mamba slot allocator for both running states and cached dense state checkpoints, plus a separately allocated, fixed-capacity DASC compressed pool. Dense cached slots can therefore be reclaimed for active computation. Radix nodes record the representation tier, slot identifier, and lock reference count; dense and compressed cached nodes share one global Mamba LRU order. \texttt{UnifiedRadixCache} unifies cache metadata and lifecycle management, not raw HBM allocation: this integration does not place KV, dense Mamba state, and compressed DASC state in one \texttt{UnifiedKVPool}. The unified-HBM construction path still sets \texttt{mamba\_ckpt\_pool=None}.

\paragraph{Insertion and reuse.} DASC precomputes selection masks, ragged offsets, and optional TP placement maps once and reuses them for state checkpoint packing and restoration. Runtime insertion is compressed-first. When a compressed slot is available, the runtime packs the selected units and releases the dense source only after the transfer completes and ownership permits release. When the compressed pool is full, an eligible completed request transfers its dense active-slot ownership to Radix without an additional copy. A dense hit uses an exact batched copy-on-write into active state and bypasses DASC recovery. A compressed hit restores retained global units; \norecon{} zero-fills omitted units, whereas \recon{} refreshes them using the bounded scratch re-prefill in Section~\ref{sec:load}. The remaining request suffix then executes normally. Placement collectives occur only at compressed state checkpoint store/load boundaries; the load transfer inverts the store mapping, and token-wise recurrence remains rank-local.

\paragraph{Pressure-driven reclamation.} When free native slots fall below the reclamation watermark, the runtime considers only cached nodes that are unlocked and unreferenced by running requests. If compressed space is available, it demotes the oldest eligible dense state checkpoint. If both pools are full and the oldest eligible global-LRU node is dense, it evicts that node directly. If that node is compressed, the runtime evicts it and uses the freed compressed slot to demote the oldest eligible dense node. Demotion preserves the node's original LRU position: changing representation is not a cache access. Evicting a compressed node alone cannot free a native slot. With no eligible dense source, this reclamation path cannot satisfy active-state demand; scheduling must wait or apply backpressure. KV eviction is useful only where the allocator actually permits the released memory to satisfy that allocation.

\paragraph{Transactional demotion.} Demotion pins the Mamba component, packs on a background CUDA stream, and records a completion event. Until completion, the dense source remains valid and cannot be reused. The runtime commits the new tier and slot and releases the source only after the event completes and concurrent cache-hit locks clear. Logically, a node passes through dense, demoting, compressed, and absent states. Victim selection and slot mappings must agree across TP ranks, and batched reclamation must preserve this ordering. A dense-victim index may accelerate selection, but it must use the same recency ordering rather than define a second eviction policy.

\paragraph{Memory boundary and costs.} The fixed compressed allocation cannot be borrowed directly for KV or running states. Dense spill retains elasticity within the native Mamba allocation, while watermark demotion can reduce request-path stalls without eliminating pack, transfer, and event costs. A total-HBM comparison must charge native-state capacity, compressed storage, staging/scratch buffers, and KV allocations to the same budget in both systems; compressed capacity must not be added on top of an unchanged native budget. The WildChat serving experiments use this integration; Appendix~\ref{app:stats} distinguishes allocated pool capacity from runtime residency.

\section{Evaluation protocol}
\label{app:stats}
\subsection{Shared system and state checkpoint accounting}
All formal experiments run in SGLang. Kimi-Linear and Qwen3-Next use eight Hopper-architecture GPUs with tensor parallelism (TP) 8; their model execution and weights use BF16 without model quantization. Recurrent state checkpoints use SGLang's default mixed precision: FP32 temporal state and BF16 convolution state. The evaluated configuration uses \texttt{mem-fraction-static}=0.80, the overlap scheduler disabled, and cache telemetry enabled. WildChat uses the two-tier state checkpoint integration in Appendix~\ref{app:pool-management}: native active states and dense spill share a Mamba slot allocator, while a separate fixed pool stores ragged compressed state checkpoints. Both cached representations share one global Mamba LRU. Dense retains full mixed-precision state checkpoints without DASC packing. Across arms, model weights, tokenizer and chat template, frozen input IDs, prompt hashes, request order, and sampling seeds are identical.

Matched-HBM serving charges both resident tiers and temporary state buffers to the allocated state-memory budget. With unchanged KV and model allocations, the per-rank comparison satisfies
\[
 B_{\mathrm{native}}^{\mathrm{Dense}}+B_{\mathrm{tmp}}^{\mathrm{Dense}}
 =B_{\mathrm{native}}^{\mathrm{DASC}}+B_{\mathrm{compressed}}^{\mathrm{DASC}}+B_{\mathrm{tmp}}^{\mathrm{DASC}}.
\]
Here native memory includes running states and dense cached state checkpoints; temporary memory includes staging and scratch allocations. The fixed compressed pool is not additional memory on top of an unchanged Dense native allocation. Physical state checkpoint size includes FP32 temporal state and uncompressed BF16 convolution state, and TP capacity is limited by the bottleneck rank.

For a separately allocated pool of per-rank budget $B_{\mathrm{pool}}$ and physical bytes per slot $b_r^{\mathrm{full}}$, its usable slot count is $\lfloor B_{\mathrm{pool}}/\max_r b_r^{\mathrm{full}}\rfloor-1$, reserving allocator slot~0. This is an allocation-capacity calculation, not a count of all resident prefixes: dense spill can occupy otherwise free native slots, subject to active-request demand. Compression ratios describe the state checkpoint representation or configured capacity, not an equal reduction in total process HBM. Full-attention KV remains unchanged and is excluded from these compression ratios.

\subsection{Quality-evaluation protocols}

\paragraph{Compressed-only state checkpoint caching.} In DASC and Random accuracy runs, all cached recurrent state checkpoints reside in the compressed pool; dense checkpoint spill is disabled. Native dense slots are used for active computation, not checkpoint caching, so a state checkpoint hit cannot bypass compression through a dense cached copy. The Dense baseline retains full state checkpoints. This restriction does not imply that every replay request is a cache hit.

\paragraph{Shared warmup--replay accounting.} RULER and Table~\ref{tab:reasoning} first issue an unscored warm request for each evaluated prefix and then issue measured replay requests. Warmup outputs and tokens are discarded. Accuracy or F1 includes every measured replay output rather than conditioning on a successful cache hit; Table~\ref{tab:reasoning} reports Hit pooled across all compared arms in each row, whose hit rates are similar,
$H=\sum_i C_i/\sum_i(C_i+N_i)$, where $i$ ranges over measured replay requests across these arms and $C_i$/$N_i$ count cached/new input tokens. Replays are averaged within each unique instance or problem, which remains the statistical unit, and paired arms share prompts, warmup/replay order, seeds, and slot counts.

\paragraph{RULER.} Each subtask--length--arm setting uses $N=30$ unique instances and three measured replay rounds. The complete per-subtask sweeps are in Tables~\ref{tab:ruler-kda-extra} and~\ref{tab:ruler-gdn-full}.

\paragraph{Controlled suffix-refresh accuracy.}
Figure~\ref{fig:kda-gdn-overview}(d,h) follows the shared warmup--replay accounting with additional cache-residency controls. Panel~(d) reports Kimi-Linear replay accuracy for five reasoning benchmarks at $\wmax=128$; panel~(h) reports the Kimi-Linear MMLU-Pro sweep over $\wmax\in\{16,64,128,512\}$. Panel~(l) pairs Kimi-Linear MMLU-Pro replay accuracy with WildChat TTFT from a separate workload, so it is a cross-workload operating-point view rather than a single-run metric.

The MMLU-Pro protocol fixes 512 state checkpoint slots, 192 questions per block, concurrency 256, and prompt-only state checkpointing, so Dense, \norecon{}, and \recon{} retain the same prompt boundaries and continuation states cannot evict them. Each block is flushed, filled by one warm request per question (\texttt{try\_id}=0), checked for at least 272 available slots, and replayed seven times per question (\texttt{try\_id}=1,\ldots,7). Each block must have zero replay eviction, token hit $>0.70$, and at least 95\% loaded requests; each complete arm must have at least 99\% aggregate loaded requests.

Accuracy includes all measured replay outputs, regardless of cache-hit status, and excludes warmup outputs. Paired deltas are computed on the same full set of measured $(\text{question},\text{try})$ pairs across arms, without filtering to successfully loaded pairs. Kimi-Linear uses temperature 1.0 and top-$p$ 1.0 and evaluates all 12,032 MMLU-Pro questions with one warm and seven measured replays per question.

\paragraph{Five-benchmark end-task suite.} Table~\ref{tab:reasoning} uses the shared two-stage schedule but not the additional block-residency qualification above; its measured replay requests run through the standard concurrent serving path. We evaluate all unique problems in AIME~2026 ($N=30$), HMMT~2026 Feb ($N=33$), IMOAnswerBench ($N=400$), GPQA-Diamond ($N=198$), and MMLU-Pro ($N=12{,}032$). Each problem receives one unscored warmup followed by $31$, $31$, $3$, $15$, or $7$ measured replays for AIME, HMMT, IMOAnswerBench, GPQA-Diamond, or MMLU-Pro, respectively ($32$, $32$, $4$, $16$, or $8$ total generations). Only measured replay outputs contribute to accuracy. Kimi-K3 evaluates AIME, HMMT, IMOAnswerBench, and MMLU-Pro with the same semantics, and LoCoMo under the protocol below. Its DASC entries are point estimates and its Random entries are five-mask mean$\pm$sample SD. All three models use the same count-matched Random mask seeds $\{1234,2027,3407,4519,5683\}$.

\paragraph{LoCoMo.} Each arm evaluates the same $1{,}986$ questions using one measured greedy replay per question after unscored cache population of the shared serialized conversation prefix. We uniformly apply ordinary token F1, without category-specific scoring, and average across questions; each question is the statistical unit.

\subsection{WildChat matched-HBM performance protocol}
\label{app:serving-protocol}
Serving performance is evaluated using natural conversations from WildChat-1M~\citep{zhao2024wildchat}. Each model uses a deterministic, tokenizer-specific trace containing 256 four-turn conversations, or 1,024 requests. Initial prompts contain $1$k--$8$k tokens, final contexts are capped at $16$k tokens, and decode length follows the recorded WildChat assistant-response length, clipped to 1,024 tokens. No synthetic prefix concatenation is used. Unlike the compressed-only accuracy runs, DASC serving enables compressed-first insertion, dense spill, and global-LRU demotion as described in Appendix~\ref{app:pool-management}.

The matrix compares Dense with \norecon{}/\recon{}, with and without TP-balanced placement, at $\wmax\in\{16,64,128,512\}$ for Kimi-Linear and Qwen3-Next and $\wmax\in\{64,128,512\}$ for Kimi-K3. Within a model, all arms use the same allocated state-memory budget, including native slots, compressed storage, and temporary buffers. Full-attention KV allocation is unchanged and excluded from the state checkpoint compression ratio. All three models use the same five run seeds. For each seed, TTFT and E2E latency are averaged over all 1,024 requests rather than conditioned on cache hits; Figure~\ref{fig:kda-gdn-overview} reports the mean across seeds. The measurements therefore include misses, state checkpoint lookup and load, prefill, decoding, and DASC runtime overhead.

\paragraph{Kimi-K3 under increased queueing pressure.}
The main serving experiments use client concurrency 256 and a server concurrency limit of 32, with no appreciable queueing. We additionally evaluate the selected Kimi-K3 \norecon{} configuration at $\wmax=64$ with Off, keeping client concurrency at 256 and lowering the server concurrency limit to 16 or 8. These settings increase queueing pressure rather than the configured client concurrency. Each setting is evaluated over five seeds. Table~\ref{tab:k3-queueing} reports mean and P95 TTFT and E2E latency. Each statistic is computed separately within each seed and then averaged across the five seeds; P95 is not computed from pooled requests across seeds. At server limits 16/8, mean TTFT decreases by $15.70/11.26\%$ and mean E2E latency by $11.25/9.22\%$, respectively. Both P95 metrics also improve, supporting the selected configuration under the tested queueing conditions without implying optimality at every load.

\begin{table}[!htbp]
\centering
\caption{Kimi-K3 latency under increased queueing pressure (seconds; lower is better). Client concurrency is 256; each setting uses five seeds. DASC uses \norecon{} at $\wmax=64$ with Off.}
\label{tab:k3-queueing}
\small
\setlength{\tabcolsep}{5pt}
\renewcommand{\arraystretch}{1.0}
\begin{tabular*}{\linewidth}{@{\extracolsep{\fill}}clrrrr@{}}
\toprule
\shortstack{Server\\limit} & Method & \multicolumn{2}{c}{TTFT (s)} & \multicolumn{2}{c}{E2E latency (s)} \\
\cmidrule(lr){3-4}\cmidrule(lr){5-6}
 & & Mean & P95 & Mean & P95 \\
\midrule
\multirow{2}{*}{16} & Dense & 18.392 & 50.946 & 27.238 & 62.450 \\
 & \norecon{} & 15.505 & 41.964 & 24.174 & 53.157 \\
\midrule
\multirow{2}{*}{8} & Dense & 54.453 & 93.929 & 76.547 & 131.924 \\
 & \norecon{} & 48.323 & 88.864 & 69.488 & 126.528 \\
\bottomrule
\end{tabular*}
\end{table}

\FloatBarrier

\section{Complete quality results}
\label{app:ruler-full}
Tables~\ref{tab:ruler-kda-extra} and~\ref{tab:ruler-gdn-full} report the complete Kimi-Linear and Qwen3-Next RULER sweeps, respectively.

\begin{table*}[!htbp]
\caption{Full replay-only Kimi-Linear (KDA) RULER sweep ($N=30$ per subtask--length; replay only).}
\label{tab:ruler-kda-extra}
\centering
\vspace{-0.2em}
\renewcommand{\arraystretch}{0.74}
\resizebox{\textwidth}{!}{%
\begin{tabular}{ll l ccc!{\vrule width 0.3pt}ccc!{\vrule width 0.3pt}cccccccc}
\toprule
& \bf Arm & \bf Len & S1 & S2 & S3 & MK1 & MK2 & MK3 & MQ & MV & CWE & FWE & QA-H & QA-S & VT & \bf Avg \\
\midrule
\multirow{3}{*}{\rotatebox{90}{baseline}} & \multirow{3}{*}{dense} & 4k & 1.00 & 1.00 & 1.00 & 1.00 & 1.00 & 1.00 & 1.00 & 0.99 & 1.00 & 0.89 & 0.57 & 0.90 & 1.00 & 0.95 \\
 &  & 8k & 1.00 & 1.00 & 1.00 & 1.00 & 1.00 & 1.00 & 1.00 & 1.00 & 1.00 & 0.94 & 0.62 & 0.86 & 1.00 & 0.96 \\
 &  & 16k & 1.00 & 1.00 & 1.00 & 1.00 & 1.00 & 1.00 & 1.00 & 0.99 & 1.00 & 0.94 & 0.57 & 0.93 & 0.98 & 0.95 \\
\cmidrule(l){2-17}
\multirow{15}{*}{\rotatebox{90}{\norecon{}}} & \multirow{3}{*}{$W_{\max}$16} & 4k & 1.00 & 1.00 & 1.00 & 1.00 & 1.00 & 1.00 & 1.00 & 0.99 & 1.00 & 0.89 & 0.59 & 0.90 & 1.00 & 0.95 \\
 &  & 8k & 1.00 & 1.00 & 1.00 & 1.00 & 1.00 & 1.00 & 1.00 & 1.00 & 1.00 & 0.94 & 0.59 & 0.87 & 1.00 & 0.95 \\
 &  & 16k & 1.00 & 1.00 & 1.00 & 1.00 & 1.00 & 1.00 & 1.00 & 1.00 & 1.00 & 0.94 & 0.56 & 0.88 & 0.99 & 0.95 \\
\cmidrule(l){2-17}
 & \multirow{3}{*}{$W_{\max}$64} & 4k & 1.00 & 1.00 & 1.00 & 1.00 & 1.00 & 1.00 & 1.00 & 0.96 & 1.00 & 0.89 & 0.58 & 0.92 & 1.00 & 0.95 \\
 &  & 8k & 1.00 & 1.00 & 1.00 & 1.00 & 1.00 & 1.00 & 1.00 & 0.97 & 0.99 & 0.93 & 0.59 & 0.78 & 1.00 & 0.94 \\
 &  & 16k & 1.00 & 1.00 & 1.00 & 1.00 & 1.00 & 1.00 & 1.00 & 0.98 & 0.99 & 0.95 & 0.54 & 0.84 & 0.99 & 0.95 \\
\cmidrule(l){2-17}
 & \multirow{3}{*}{$W_{\max}$128} & 4k & 1.00 & 1.00 & 0.98 & 1.00 & 1.00 & 1.00 & 1.00 & 0.97 & 0.89 & 0.89 & 0.53 & 0.91 & 1.00 & 0.94 \\
 &  & 8k & 1.00 & 1.00 & 0.97 & 1.00 & 1.00 & 1.00 & 1.00 & 0.95 & 0.91 & 0.93 & 0.57 & 0.84 & 1.00 & 0.94 \\
 &  & 16k & 1.00 & 1.00 & 0.97 & 0.98 & 1.00 & 1.00 & 1.00 & 0.97 & 0.82 & 0.94 & 0.53 & 0.92 & 1.00 & 0.93 \\
\cmidrule(l){2-17}
 & \multirow{3}{*}{$W_{\max}$512} & 4k & 1.00 & 1.00 & 1.00 & 0.99 & 1.00 & 1.00 & 1.00 & 0.89 & 0.85 & 0.86 & 0.47 & 0.87 & 0.99 & 0.92 \\
 &  & 8k & 1.00 & 1.00 & 1.00 & 0.96 & 1.00 & 0.97 & 1.00 & 0.86 & 0.70 & 0.91 & 0.62 & 0.84 & 0.98 & 0.91 \\
 &  & 16k & 1.00 & 1.00 & 1.00 & 0.88 & 0.97 & 0.98 & 1.00 & 0.89 & 0.66 & 0.94 & 0.57 & 0.84 & 0.96 & 0.90 \\
\cmidrule(l){2-17}
 & \multirow{3}{*}{$W_{\max}$1024} & 4k & 1.00 & 0.99 & 1.00 & 1.00 & 1.00 & 1.00 & 0.98 & 0.87 & 0.75 & 0.86 & 0.53 & 0.86 & 1.00 & 0.91 \\
 &  & 8k & 1.00 & 1.00 & 1.00 & 0.96 & 0.97 & 0.94 & 1.00 & 0.85 & 0.76 & 0.92 & 0.59 & 0.80 & 0.96 & 0.90 \\
 &  & 16k & 1.00 & 0.98 & 1.00 & 0.82 & 0.94 & 0.98 & 1.00 & 0.83 & 0.72 & 0.94 & 0.52 & 0.79 & 0.92 & 0.88 \\
\midrule
\multirow{15}{*}{\rotatebox{90}{\recon{}}} & \multirow{3}{*}{$W_{\max}$16} & 4k & 1.00 & 1.00 & 1.00 & 1.00 & 1.00 & 1.00 & 1.00 & 1.00 & 1.00 & 0.89 & 0.59 & 0.90 & 1.00 & 0.95 \\
 &  & 8k & 1.00 & 1.00 & 1.00 & 1.00 & 1.00 & 1.00 & 1.00 & 1.00 & 1.00 & 0.94 & 0.57 & 0.86 & 1.00 & 0.95 \\
 &  & 16k & 1.00 & 1.00 & 1.00 & 1.00 & 1.00 & 1.00 & 1.00 & 0.99 & 1.00 & 0.94 & 0.57 & 0.87 & 0.99 & 0.95 \\
\cmidrule(l){2-17}
 & \multirow{3}{*}{$W_{\max}$64} & 4k & 1.00 & 1.00 & 1.00 & 0.98 & 1.00 & 1.00 & 1.00 & 0.96 & 1.00 & 0.89 & 0.57 & 0.90 & 1.00 & 0.95 \\
 &  & 8k & 1.00 & 1.00 & 1.00 & 1.00 & 1.00 & 1.00 & 1.00 & 0.95 & 1.00 & 0.93 & 0.60 & 0.82 & 0.99 & 0.95 \\
 &  & 16k & 1.00 & 1.00 & 1.00 & 1.00 & 1.00 & 1.00 & 1.00 & 0.97 & 0.91 & 0.94 & 0.58 & 0.87 & 0.96 & 0.94 \\
\cmidrule(l){2-17}
 & \multirow{3}{*}{$W_{\max}$128} & 4k & 1.00 & 1.00 & 1.00 & 0.99 & 1.00 & 1.00 & 1.00 & 0.91 & 0.95 & 0.88 & 0.56 & 0.90 & 1.00 & 0.94 \\
 &  & 8k & 1.00 & 1.00 & 1.00 & 0.97 & 1.00 & 1.00 & 1.00 & 0.88 & 0.95 & 0.93 & 0.60 & 0.84 & 0.98 & 0.93 \\
 &  & 16k & 1.00 & 1.00 & 1.00 & 1.00 & 1.00 & 0.97 & 1.00 & 0.85 & 0.87 & 0.94 & 0.58 & 0.87 & 0.99 & 0.93 \\
\cmidrule(l){2-17}
 & \multirow{3}{*}{$W_{\max}$512} & 4k & 1.00 & 1.00 & 1.00 & 1.00 & 1.00 & 1.00 & 1.00 & 0.89 & 0.96 & 0.88 & 0.57 & 0.90 & 1.00 & 0.94 \\
 &  & 8k & 1.00 & 1.00 & 1.00 & 0.94 & 0.99 & 1.00 & 1.00 & 0.82 & 0.98 & 0.93 & 0.59 & 0.84 & 0.99 & 0.93 \\
 &  & 16k & 1.00 & 1.00 & 1.00 & 0.96 & 0.98 & 1.00 & 1.00 & 0.84 & 0.81 & 0.94 & 0.64 & 0.88 & 0.99 & 0.93 \\
\cmidrule(l){2-17}
 & \multirow{3}{*}{$W_{\max}$1024} & 4k & 1.00 & 1.00 & 1.00 & 0.98 & 1.00 & 1.00 & 1.00 & 0.96 & 0.97 & 0.88 & 0.57 & 0.90 & 1.00 & 0.94 \\
 &  & 8k & 1.00 & 1.00 & 1.00 & 0.94 & 0.99 & 1.00 & 1.00 & 0.86 & 0.97 & 0.93 & 0.60 & 0.82 & 1.00 & 0.93 \\
 &  & 16k & 1.00 & 1.00 & 1.00 & 1.00 & 1.00 & 1.00 & 1.00 & 0.87 & 0.96 & 0.94 & 0.64 & 0.88 & 1.00 & 0.95 \\
\bottomrule
\end{tabular}%
}

\par\medskip
\caption{Full replay-only Qwen3-Next (GDN) RULER sweep ($N=30$ per subtask--length).}
\label{tab:ruler-gdn-full}
\centering
\vspace{-0.2em}
\renewcommand{\arraystretch}{0.74}
\resizebox{\textwidth}{!}{%
\begin{tabular}{ll l ccc!{\vrule width 0.3pt}ccc!{\vrule width 0.3pt}cccccccc}
\toprule
& \bf Arm & \bf Len & S1 & S2 & S3 & MK1 & MK2 & MK3 & MQ & MV & CWE & FWE & QA-H & QA-S & VT & \bf Avg \\
\midrule
\multirow{3}{*}{\rotatebox{90}{baseline}} & \multirow{3}{*}{dense} & 4k & 1.00 & 1.00 & 1.00 & 1.00 & 1.00 & 1.00 & 0.74 & 0.97 & 0.39 & 0.98 & 0.57 & 0.92 & 0.37 & 0.84 \\
 &  & 8k & 1.00 & 1.00 & 1.00 & 1.00 & 1.00 & 1.00 & 0.75 & 0.84 & 0.35 & 0.97 & 0.62 & 0.82 & 0.40 & 0.83 \\
 &  & 16k & 1.00 & 1.00 & 1.00 & 1.00 & 1.00 & 1.00 & 0.75 & 0.73 & 0.29 & 0.93 & 0.51 & 0.83 & 0.41 & 0.80 \\
\cmidrule(l){2-17}
\multirow{15}{*}{\rotatebox{90}{\norecon{}}} & \multirow{3}{*}{$W_{\max}$16} & 4k & 1.00 & 1.00 & 1.00 & 1.00 & 1.00 & 1.00 & 0.74 & 0.98 & 0.38 & 0.97 & 0.57 & 0.92 & 0.38 & 0.84 \\
 &  & 8k & 1.00 & 1.00 & 1.00 & 1.00 & 1.00 & 1.00 & 0.75 & 0.86 & 0.34 & 0.97 & 0.60 & 0.84 & 0.40 & 0.83 \\
 &  & 16k & 1.00 & 1.00 & 1.00 & 1.00 & 1.00 & 1.00 & 0.75 & 0.74 & 0.31 & 0.93 & 0.53 & 0.84 & 0.42 & 0.81 \\
\cmidrule(l){2-17}
 & \multirow{3}{*}{$W_{\max}$64} & 4k & 1.00 & 1.00 & 1.00 & 1.00 & 1.00 & 1.00 & 0.74 & 0.97 & 0.36 & 0.97 & 0.54 & 0.92 & 0.36 & 0.84 \\
 &  & 8k & 1.00 & 1.00 & 1.00 & 1.00 & 1.00 & 1.00 & 0.75 & 0.85 & 0.37 & 0.97 & 0.61 & 0.82 & 0.39 & 0.83 \\
 &  & 16k & 1.00 & 1.00 & 1.00 & 1.00 & 1.00 & 1.00 & 0.75 & 0.74 & 0.28 & 0.93 & 0.50 & 0.82 & 0.42 & 0.80 \\
\cmidrule(l){2-17}
 & \multirow{3}{*}{$W_{\max}$128} & 4k & 1.00 & 1.00 & 1.00 & 1.00 & 1.00 & 1.00 & 0.75 & 0.94 & 0.40 & 0.97 & 0.57 & 0.92 & 0.38 & 0.84 \\
 &  & 8k & 1.00 & 1.00 & 1.00 & 1.00 & 1.00 & 0.99 & 0.75 & 0.85 & 0.40 & 0.97 & 0.61 & 0.82 & 0.41 & 0.83 \\
 &  & 16k & 1.00 & 1.00 & 1.00 & 1.00 & 1.00 & 0.94 & 0.75 & 0.73 & 0.30 & 0.93 & 0.53 & 0.84 & 0.43 & 0.80 \\
\cmidrule(l){2-17}
 & \multirow{3}{*}{$W_{\max}$512} & 4k & 1.00 & 1.00 & 1.00 & 1.00 & 1.00 & 0.97 & 0.75 & 0.96 & 0.30 & 0.97 & 0.54 & 0.92 & 0.36 & 0.83 \\
 &  & 8k & 1.00 & 1.00 & 1.00 & 1.00 & 1.00 & 0.96 & 0.75 & 0.86 & 0.30 & 0.97 & 0.58 & 0.82 & 0.39 & 0.82 \\
 &  & 16k & 1.00 & 1.00 & 1.00 & 1.00 & 1.00 & 0.89 & 0.75 & 0.75 & 0.24 & 0.93 & 0.53 & 0.87 & 0.40 & 0.80 \\
\cmidrule(l){2-17}
 & \multirow{3}{*}{$W_{\max}$1024} & 4k & 1.00 & 1.00 & 1.00 & 1.00 & 1.00 & 1.00 & 0.73 & 0.94 & 0.35 & 0.97 & 0.56 & 0.92 & 0.35 & 0.83 \\
 &  & 8k & 1.00 & 1.00 & 1.00 & 1.00 & 1.00 & 0.97 & 0.75 & 0.86 & 0.35 & 0.97 & 0.62 & 0.84 & 0.40 & 0.83 \\
 &  & 16k & 1.00 & 1.00 & 1.00 & 1.00 & 1.00 & 0.87 & 0.75 & 0.75 & 0.27 & 0.93 & 0.49 & 0.84 & 0.38 & 0.79 \\
\midrule
\multirow{15}{*}{\rotatebox{90}{\recon{}}} & \multirow{3}{*}{$W_{\max}$16} & 4k & 1.00 & 1.00 & 1.00 & 1.00 & 1.00 & 1.00 & 0.75 & 0.96 & 0.20 & 0.97 & 0.56 & 0.92 & 0.37 & 0.83 \\
 &  & 8k & 1.00 & 1.00 & 1.00 & 1.00 & 1.00 & 1.00 & 0.75 & 0.85 & 0.54 & 0.97 & 0.60 & 0.84 & 0.39 & 0.84 \\
 &  & 16k & 1.00 & 1.00 & 1.00 & 1.00 & 1.00 & 1.00 & 0.75 & 0.75 & 0.43 & 0.93 & 0.51 & 0.84 & 0.42 & 0.82 \\
\cmidrule(l){2-17}
 & \multirow{3}{*}{$W_{\max}$64} & 4k & 1.00 & 1.00 & 1.00 & 1.00 & 1.00 & 1.00 & 0.74 & 0.94 & 0.23 & 0.98 & 0.57 & 0.92 & 0.38 & 0.83 \\
 &  & 8k & 1.00 & 1.00 & 1.00 & 1.00 & 1.00 & 1.00 & 0.75 & 0.86 & 0.53 & 0.97 & 0.61 & 0.82 & 0.40 & 0.84 \\
 &  & 16k & 1.00 & 1.00 & 1.00 & 1.00 & 1.00 & 1.00 & 0.75 & 0.72 & 0.34 & 0.93 & 0.51 & 0.83 & 0.42 & 0.81 \\
\cmidrule(l){2-17}
 & \multirow{3}{*}{$W_{\max}$128} & 4k & 1.00 & 1.00 & 1.00 & 1.00 & 1.00 & 1.00 & 0.74 & 0.95 & 0.24 & 0.97 & 0.54 & 0.92 & 0.38 & 0.83 \\
 &  & 8k & 1.00 & 1.00 & 1.00 & 1.00 & 1.00 & 1.00 & 0.75 & 0.85 & 0.54 & 0.97 & 0.58 & 0.82 & 0.40 & 0.84 \\
 &  & 16k & 1.00 & 1.00 & 1.00 & 1.00 & 1.00 & 1.00 & 0.75 & 0.76 & 0.37 & 0.93 & 0.50 & 0.84 & 0.41 & 0.81 \\
\cmidrule(l){2-17}
 & \multirow{3}{*}{$W_{\max}$512} & 4k & 1.00 & 1.00 & 1.00 & 1.00 & 1.00 & 1.00 & 0.75 & 0.95 & 0.21 & 0.97 & 0.54 & 0.92 & 0.38 & 0.82 \\
 &  & 8k & 1.00 & 1.00 & 1.00 & 1.00 & 1.00 & 0.98 & 0.75 & 0.86 & 0.54 & 0.97 & 0.60 & 0.82 & 0.39 & 0.84 \\
 &  & 16k & 1.00 & 1.00 & 1.00 & 1.00 & 0.98 & 0.90 & 0.75 & 0.74 & 0.43 & 0.93 & 0.53 & 0.84 & 0.43 & 0.81 \\
\cmidrule(l){2-17}
 & \multirow{3}{*}{$W_{\max}$1024} & 4k & 1.00 & 1.00 & 1.00 & 1.00 & 1.00 & 0.99 & 0.73 & 0.94 & 0.24 & 0.97 & 0.57 & 0.92 & 0.39 & 0.83 \\
 &  & 8k & 1.00 & 1.00 & 1.00 & 1.00 & 1.00 & 0.97 & 0.75 & 0.83 & 0.43 & 0.97 & 0.61 & 0.82 & 0.40 & 0.83 \\
 &  & 16k & 1.00 & 1.00 & 1.00 & 1.00 & 0.98 & 0.83 & 0.75 & 0.76 & 0.35 & 0.93 & 0.50 & 0.83 & 0.48 & 0.80 \\
\bottomrule
\end{tabular}%
}

\end{table*}

\FloatBarrier

\section{State quantization and \dasc{} composition}

\subsection{INT8 state checkpoint protocol}
\label{app:int8-protocol}

This state-quantization study evaluates Kimi-Linear only. Its unquantized Dense and \dasc{} arms use SGLang's default mixed-precision recurrent state checkpoint: FP32 temporal state and BF16 convolution state. Model weights and execution remain BF16 and are not quantized. For INT8 arms, we quantize only the cached FP32 temporal recurrent state $S$; convolution state remains BF16 and the active temporal state used by recurrence remains FP32. Quantization is symmetric signed INT8 over $[-127,127]$, without a zero point. For a state tensor $S\in\mathbb{R}^{L\times H\times d_v\times d_k}$, one scale is used for each $(l,h,k)$ tuple and shared over $d_v$:
\begin{equation}
a_{l,h,k}=\max_v |S_{l,h,v,k}|,\qquad
s_{l,h,k}=\frac{\max(a_{l,h,k},10^{-8})}{127},\qquad
q=\operatorname{clip}\!\left(\operatorname{round}(S/s),-127,127\right).
\end{equation}
The maximum, division, and rounding are computed in FP32; $q$ is stored as INT8 and $s$ in the temporal source dtype (FP32 here). On a cache hit, the state checkpoint is dequantized once as $\hat S=q\,s$ into a fresh active FP32 temporal-state slot; subsequent recurrence and decoding use native precision. Thus state checkpoints are quantized once at insertion and dequantized once on reuse, rather than repeatedly quantized during recurrence. Conv1D window states remain BF16 because their footprint is small. Reported HBM includes the INT8 payload, FP32 scales, BF16 convolution windows, and the reserved allocator slot. These overheads produce the measured $3.88\times$ quantization-only full state checkpoint compression rather than the ideal temporal-payload $4\times$.

\subsection{End-task composition protocol}
\label{app:int8-dasc-e2e}

For Kimi-Linear, we compare the dense mixed-precision reference with ragged \norecon{} at $\wmax=16$, storing the selected temporal state in the symmetric INT8 format defined above. On a hit, the state checkpoint is dequantized once into the active FP32 temporal state before recurrence continues. The composition provides $8.11\times$ state checkpoint capacity relative to the dense mixed-precision state checkpoint. Full-attention KV remains unchanged. We use the same warmup--replay protocol, complete problem sets, replay counts, prompts, and sampling seeds as Table~\ref{tab:reasoning}; LoCoMo is not included in this composition experiment.

\FloatBarrier

\section{Limitations}
\label{sec:limits}

Our evaluation focuses on hybrid architectures; applicability to purely linear-attention or state-space models remains to be studied. We have not evaluated serving at tensor-parallel degrees larger than eight, so scalability beyond TP8 remains unverified.

\end{document}